\documentclass[10pt,twocolumn,letterpaper]{article}

\usepackage{times}
\usepackage{epsfig}
\usepackage{graphicx}
\usepackage{amsmath}
\usepackage{amssymb}
\usepackage{amsthm}
\usepackage{booktabs}
\usepackage{multirow}
\usepackage{graphicx}
\usepackage{algorithm}
\usepackage{algorithmic}
\usepackage{xcolor}
\usepackage{cite}
\usepackage{authblk}
\usepackage{appendix}
\usepackage[normalem]{ulem}

\newcommand{\coolname}{$\texttt{AHL}$\xspace}
\usepackage[pagenumbers]{cvpr}

\definecolor{cvprblue}{rgb}{0.21,0.49,0.74}
\usepackage[pagebackref,breaklinks,colorlinks,citecolor=cvprblue]{hyperref}

\title{Anomaly Heterogeneity Learning for Open-set Supervised Anomaly Detection}

\author{Jiawen Zhu\textsuperscript{1}}
\author{Choubo Ding\textsuperscript{2}}
\author{Yu Tian\textsuperscript{3}}
\author{Guansong Pang\textsuperscript{1}\thanks{Corresponding author: G. Pang (\tt\small gspang@smu.edu.sg)}}
\affil{\textsuperscript{1}School of Computing and Information Systems, Singapore Management University \\ \textsuperscript{2}Australian Institute for Machine Learning, University of Adelaide\\\textsuperscript{3}Harvard Ophthalmology AI Lab, Harvard University}

\begin{document}
\maketitle
\begin{abstract}
Open-set supervised anomaly detection (OSAD) -- a recently emerging anomaly detection area -- aims at utilizing a few samples of anomaly classes seen during training to detect unseen anomalies (\ie, samples from open-set anomaly classes), while effectively identifying the seen anomalies. 
Benefiting from the prior knowledge illustrated by the seen anomalies, current OSAD methods can often largely reduce false positive errors.
However, these methods are trained in a closed-set setting and treat the anomaly examples as from a homogeneous distribution, rendering them less effective in generalizing to unseen anomalies that can be drawn from any distribution. This paper proposes to learn heterogeneous anomaly distributions using the limited anomaly examples to address this issue. To this end, we introduce a novel approach, namely
Anomaly Heterogeneity Learning (\coolname), that simulates a diverse set of heterogeneous anomaly distributions and then utilizes them to learn a unified heterogeneous abnormality model in surrogate open-set environments. Further, \coolname is a generic framework that existing OSAD models can plug and play for enhancing their abnormality modeling.
Extensive experiments on nine real-world anomaly detection datasets show that \coolname can 1) substantially enhance different state-of-the-art OSAD models in detecting seen and unseen anomalies, and 2) effectively generalize to unseen anomalies in new domains. Code is available at \renewcommand\UrlFont{\color{blue}}\url{https://github.com/mala-lab/AHL}. 
\end{abstract}

\section{Introduction}

Anomaly detection (AD) aims at identifying data points that significantly deviate from the majority of the data. It has gained considerable attention in both academic and industry communities due to its broad applications in diverse domains such as industrial inspection, medical imaging, and scientific discovery, etc. \cite{pang2021deep}. Since it is difficult, or prohibitively costly, to collect large-scale labeled anomaly data, most existing AD approaches treat it as a one-class problem, where only normal samples are available during training~\cite{bergman2020classification, hou2021divide, park2020learning, tax2004support, yan2021learning, zaheer2020old, zavrtanik2021draem, zavrtanik2021reconstruction, salehi2021multiresolution, chen2022deep, yi2020patch, akcay2019ganomaly, schlegl2019f, deng2022anomaly, wang2021student, cao2023anomaly, tien2023revisiting, liu2023diversity, liu2023simplenet, zhang2023destseg}. However, in many applications there often exist a few accessible anomaly examples, such as defect samples identified in the past industrial inspection and tumor images of past patients. The anomaly examples offer important source of prior knowledge about abnormality, but these one-class-based approaches are unable to use them. 

\begin{figure}
    \centering
    \includegraphics[width=0.45\textwidth]{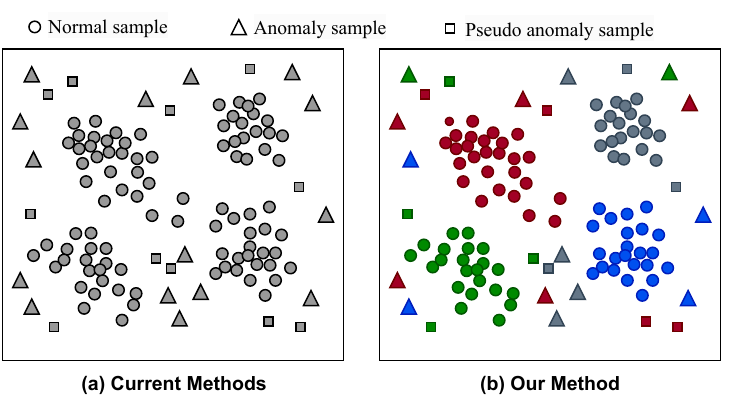}
        \caption{Current methods vs. our method \coolname, where the anomaly samples of the same color indicates that they are treated as from one data distribution. Compared to existing methods that model a \textit{homogeneous} anomaly distribution in a closed-set environment, \coolname simulates a diverse set of heterogeneous anomaly distributions (Sec. \ref{subsec:hadg}) and learns \textit{heterogeneous} abnormality from them in a surrogate open environment (Sec. \ref{subsec:cdl}).
        }
    \label{fig:intro}
    \vspace{-0.5cm}
\end{figure}

Open-set supervised AD (OSAD) is a recently emerging area that aims at utilizing those limited training anomaly data to learn generalized models for detecting \textit{unseen anomalies} (\ie, samples from open-set anomaly classes), while effectively identifying those \textit{seen anomalies} (\ie, anomalies that are similar to training anomaly examples). A number of methods have been introduced for this OSAD problem~\cite{ding2022catching,  acsintoae2022ubnormal,zhu2022towards,pang2021explainable,liu2019margin}. Benefiting from the prior knowledge illustrated by the seen anomalies, current OSAD methods can often largely reduce false positive errors. 

One issue with the current OSAD methods is that they treat the anomaly examples as from a homogeneous distribution, as shown in Fig. \ref{fig:intro}(a), which can largely restrict their performance in detecting unseen anomalies. This is because anomalies can arise from a wide range of conditions and are inherently unbounded, resulting in \textit{heterogeneous anomaly distributions} (\ie, anomalies can be drawn from very different distributions). For instance, tumor images can demonstrate different features in terms of appearance, shape, size, and position, etc., depending on the nature of the tumors. The current OSAD methods ignore those anomaly heterogeneity and often fail to detect anomalies if they are drawn from data distributions dissimilar to the seen anomalies.

To address this issue, we propose to learn heterogeneous anomaly distributions with the limited training anomaly examples. These anomalies are examples of seen anomaly classes only, which do not illustrate the distribution of every possible anomaly classes, \eg, those unseen ones, making it challenging to learn the underlying heterogeneous anomaly distributions with the limited anomaly information. This work introduces a novel framework, namely Anomaly Heterogeneity Learning (\coolname), to tackle this challenge. As illustrated in Fig. \ref{fig:intro}(b), it first simulates a variety of heterogeneous anomaly distributions by associating fine-grained distributions of normal examples with randomly selected anomaly examples. \coolname then performs a collaborative differentiable learning that synthesizes all these anomaly distributions to learn a heterogeneous abnormality model. Further, the generated anomaly data enables the training of our model in surrogate open environments, in which part of anomaly distributions are used for model training while the others are used as unseen data to validate and tune the model, resulting in better generalized models than the current methods that are trained in a closed-set setting. 

Additionally, the simulated anomaly distributions are typically of different quality. Thus, a self-supervised generalizability estimation is devised in \coolname to adaptively adjust the importance of each learned anomaly distribution during our model training.

A straightforward alternative approach to \coolname is to build an ensemble model based on a simple integration of homogeneous/heterogeneous OSAD models on the simulated heterogeneous data distributions. However, such ensembles fail to consider the commonalities and differences in the anomaly heterogeneity captured in the base models, leading to a suboptimal learning of the heterogeneity (Sec. \ref{subsec:ablation}).

Accordingly, this paper makes four main contributions.
\begin{itemize}
    \item \textbf{Framework.} We propose Anomaly Heterogeneity Learning (\coolname), a novel framework for OSAD. Unlike current methods that treat the training anomaly examples as a homogeneous distribution, \coolname learns heterogeneous anomaly distributions with these limited examples, enabling more generalized detection on unseen anomalies.
    \item \textbf{Novel Model.} We further instantiate the \coolname framework to a novel OSAD model. The model performs a collaborative differentiable learning of the anomaly heterogeneity using a diverse set of simulated heterogeneous anomaly distributions, facilitating an iterative validating and tuning of the model in surrogate open-set environments. This enables a more optimal anomaly heterogeneity learning than simple ensemble methods.
    \item \textbf{Generic.} Our model is generic, in which features and loss functions from different OSAD models can plug-and-play and gain substantially improved detection performance. 
    \item \textbf{Strong Generalization Ability.} Experiments on nine real-world AD datasets show that \coolname substantially outperforms state-of-the-art models in detecting unseen anomalies in the same-domain and cross-domain settings.
\end{itemize}

\section{Related Work}

\noindent\textbf{Unsupervised Anomaly Detection.}
Most existing AD approaches rely on unsupervised learning with anomaly-free training samples due to the difficulty of collecting large-scale anomaly observations. One-class classification methods aim to learn a compact normal data description using support vectors~\cite{tax2004support, yi2020patch, bergman2020classification,chen2022deep,ruff2020deep}. Another widely used AD approach learns to reconstruct normal data based on generative models such as Autoencoder (AE)~\cite{kingma2013auto} and Generative Adversarial Networks (GAN)~\cite{goodfellow2020generative}. These reconstruction methods rely on the assumption that anomalies are more difficult to be reconstructed than normal samples~\cite{akcay2019ganomaly, schlegl2019f, zavrtanik2021reconstruction, yan2021learning, zaheer2020old, zavrtanik2021draem, park2020learning, hou2021divide, xiang2023squid, liu2023diversity}. Other popular approaches include knowledge distillation~\cite{deng2022anomaly, bergmann2020uninformed, salehi2021multiresolution, wang2021student, cao2023anomaly, tien2023revisiting, zhang2023destseg} and self-supervised learning methods~\cite{li2021cutpaste, georgescu2021anomaly, ristea2022self, huang2022self, FOD}. 
Another related line of research is domain-adapted AD~\cite{YaoZF22, WuCFL21, LuYR020}. Methods in this line typically focus on a cross-domain setup, requiring the data from multiple relevant domains, whereas we focus on training detection models in single-domain data. One major problem with all these unsupervised AD approaches is that they do not have any prior knowledge about real anomalies, which can lead to many false positive errors~\cite{ChuK20, BozicTS21, ding2022catching, pang2021explainable,ruff2020deep,pang2021toward,pang2023deep,acsintoae2022ubnormal,zhu2022towards,liu2019margin}. 

\noindent\textbf{Towards Supervised Anomaly Detection.}
Supervised AD aims to reduce the detection errors using less costly supervision information, such as weakly-supervised information like video-level supervision to detect frame-level anomalies \cite{sultani2018real,tian2021weakly,wu2020not,wu2023vadclip,lv2023unbiased,chen2023mgfn} and a small set of anomaly examples from partially observed anomaly classes~\cite{abs-2007-01760, BozicTS21, ruff2020deep, pang2021explainable, pang2021toward, pang2023deep,zaheer2022generative,pang2023deep}. OSAD addresses the problem in the latter case.
One OSAD approach is one-class metric learning, where the limited training anomalies are treated as negative samples during the normality learning~\cite{ruff2020deep,liu2019margin,pang2018learning}. However, AD is inherently an open-set task due to the unknowingness nature of anomaly, so the limited negative samples are not sufficient to support an accurate one-class learning. Recently DevNet~\cite{pang2021explainable} introduces a one-sided anomaly-focused deviation loss to tackle this problem by imposing a prior on the anomaly scores. It also establishes an OSAD evaluation benchmark. DRA~\cite{ding2022catching} enhances DevNet via a framework that learns disentangled representations of seen, pseudo, and latent residual anomalies in order to better detect both seen and unseen anomalies. More recently, BGAD \cite{yao2023explicit} uses a decision boundary generated by a normalizing flow model to learn an anomaly-informed model. PRN \cite{zhang2023prototypical} learns residual representations across multi-scale 
feature maps using both image-level and pixel-level anomaly data. However, their implementation uses training anomaly examples from all anomaly types, different from our open-set AD settings that have unseen anomaly types in test data. UBnormal \cite{acsintoae2022ubnormal} and OpenVAD \cite{zhu2022towards} extend OSAD to video data and establish corresponding benchmarks. However, these methods often treat the training anomalies as from homogeneous distributions in a closed-set setting, which can restrict their performance in detecting unseen anomalies. Using the anomaly examples for anomaly generation or pseudo anomaly labeling \cite{zaheer2022generative,astrid2023pseudobound} is explored as another way to reduce the false positives, but it is performed in an unsupervised setting. 

\section{Anomaly Heterogeneity Learning}
\label{sec:method}

\noindent\textbf{Problem Statement}:
We assume to have a set of training images and annotations $\{\omega_i, y_i) \}_{i=1}$, where $\omega_i \in \Omega \subset \mathbb{R}^{H \times W \times C}$ denotes an image with RGB channels and $y_i \in \mathcal{Y} \subset \{0,1\}$ denotes an image-level class label, with $y_i=1$ if $\omega_i$ is abnormal and $y_i=0$ otherwise. Due to the rareness of anomaly, the labeled data is often predominantly presented by normal data. Given an existing AD model $f(\cdot)$ that can be used to extract low-dimensional image features for constructing the training feature set $\mathcal{D} = \{\mathbf{x}_i, y_i \}$, where $\mathbf{x}_i = f(\omega_i) \in \mathcal{X}$ indicates corresponding $i$-th image features,
with $\mathcal{X}_{n} = \{\mathbf{x}_1, \mathbf{x}_2, ..., \mathbf{x}_N\}$ and 
$\mathcal{X}_{a} = \{\mathbf{x}_1, \mathbf{x}_2, ..., \mathbf{x}_M\}$ ($N \gg M$) respectively denoting the feature set of normal and abnormal images, then the goal of our proposed \coolname framework is to learn an anomaly detection function $g: \mathcal{X}\longrightarrow \mathbb{R}$ that is capable of assigning higher anomaly scores to anomaly images drawn from different distributions than to the normal ones. Note that in OSAD the training anomalies $\mathcal{X}_{a}$ are from seen anomaly classes $\mathcal{S}$, which is only a subset of $\mathcal{C}$ that can contain a larger set of anomaly classes during inference, \eg, $\mathcal{S} \subset \mathcal{C}$.

\subsection{Overview of Our Approach}
\begin{figure*}
    \centering
    \includegraphics[width=\textwidth]{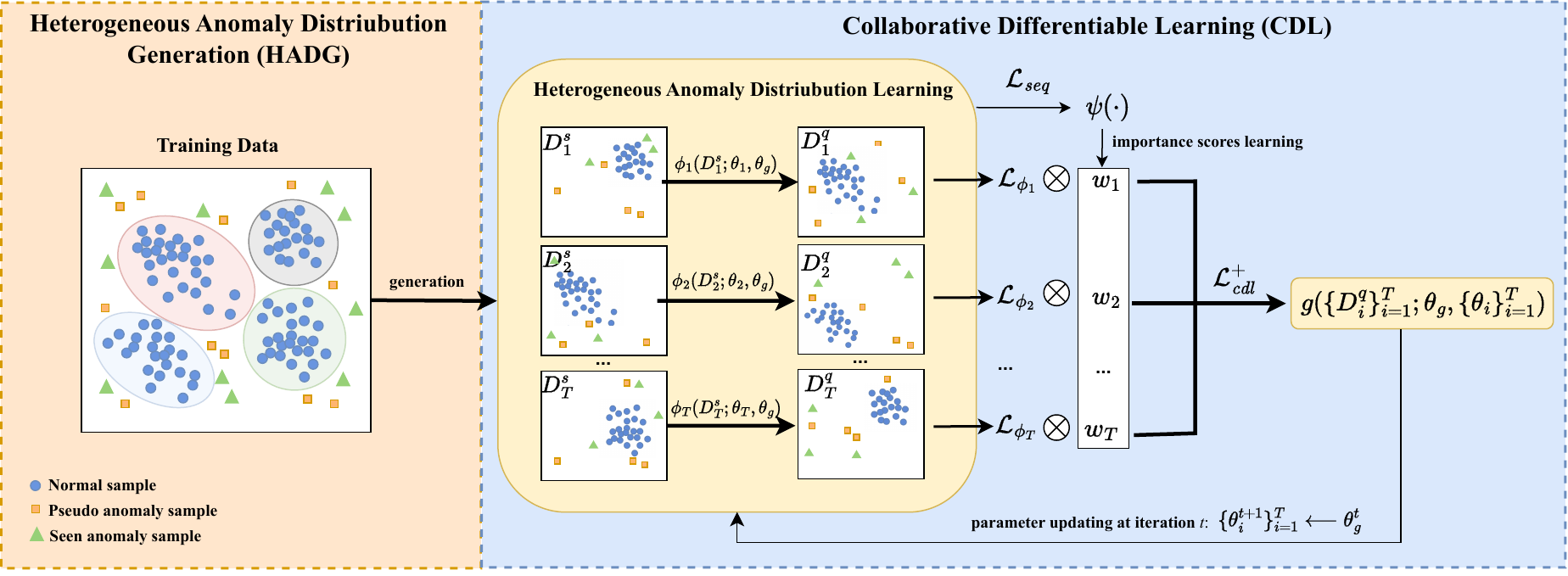}
    \caption{Overview of our approach \coolname. Its HADG component first generates $T$ heterogeneous anomaly distribution datasets from the training set, $\mathcal{T}=\{\mathcal{D}_i\}_{i=1}^T$, each of which includes a support set and open-set query set, \ie, $\mathcal{D}_i = \{\mathcal{D}^s_i, \mathcal{D}^q_i\}$. It then utilizes them to learn $T$ heterogeneous AD models $\{\phi_i\}_{i=1}^T$ in a simulated open-set environment and synthesizes these heterogeneous anomaly models into a unified AD model $g(\cdot)$ by a collaborative differential learning (CDL). Different $\phi_i$ learn anomaly distributions of various quality, so we also devise a model $\psi(\cdot)$ to assign an importance score to each $\phi_i$ to enhance the CDL component.
    }
    \label{fig:overall}
    \vspace{-0.5cm}
\end{figure*}

The key idea of our \coolname framework is to learn a unified anomaly heterogeneity model by a collaborative differentiable learning of abnormalities embedded in diverse simulated anomaly distributions. As demonstrated in Fig.~\ref{fig:overall}, \coolname consists of two main components: Heterogeneous Anomaly Distribution Generation (\textbf{HADG}) and Collaborative Differentiable Learning of the anomaly heterogeneity (\textbf{CDL}). Specifically, the HADG component simulates and generates $T$ heterogeneous distribution datasets from the training set, $\mathcal{T}=\{\mathcal{D}_i\}_{i=1}^T$, with each $\mathcal{D}_i$ containing a mixture of normal data subset and randomly sampled anomaly examples. Each $\mathcal{D}_i$ is generated in a way that represents a different anomaly distribution from the others. CDL is then designed to learn a unified heterogeneous abnormality detection model $g(\mathcal{T};\theta_g)$ that synthesizes a set of $T$ base models, denoted as $\{\phi_i(\mathcal{D}_i; \theta_i)\}_{i=1}^{T}$, where $\theta_g$ and $\theta_i$ denotes learnable weight parameters of the unified model $g$ and the base model $\phi_i$ respectively, and each $\phi_i: \mathcal{D}_i \rightarrow \mathbb{R}$ learns from one anomaly distribution for anomaly scoring. The weight parameters $\theta_g$ are collaboratively updated based on the base model weights $\{\theta_i\}_{i=1}^{T}$. Further, the effectiveness of individual base models can vary largely, so a module $\psi$ is added in CDL to increase the importance of $\theta_i$ in the collaborative weight updating if its corresponding base model $\phi_i$ is estimated to have small generalization error. During inference, only the unified heterogeneous abnormality model $g(\mathcal{T};\theta_g)$ is used for anomaly detection.

\coolname is a generic framework, in which off-the-shelf OSAD models can be easily plugged to instantiate $\phi_i$ and gain significantly improved performance.

\subsection{Heterogeneous Anomaly Distribution Data Generation}\label{subsec:hadg}

One main challenge of learning the underlying intricate abnormalities is the lack of training data that illustrate different possible anomaly distributions. Our HADG component is designed to address this challenge, in which we partition the normal examples into different clusters and associate each of the normal clusters with randomly sampled anomaly examples to create diverse anomaly distributions. The resulting distributions differ from each other in terms of normal patterns and/or abnormal patterns. Specifically, HADG generates $T$ sets of training anomaly distribution data, $\mathcal{T}=\{\mathcal{D}_i\}_{i=1}^T$, with each $\mathcal{D}_i=\mathcal{X}_{n,i} \cup \mathcal{X}_{a,i}$, where $\mathcal{X}_{n,i} \subset \mathcal{X}_{n}$ and $\mathcal{X}_{a,i} \subset \mathcal{X}_{a}$. To simulate an anomaly distribution of good quality, $\mathcal{X}_{n,i}$ should represent one main normal pattern. To this end, HADG adopts a clustering approach to partition $\mathcal{X}_n$ into $C$ clusters, and then randomly samples one of these $C$ normal clusters as $\mathcal{X}_{n,i}$. On the other hand, to ensure the diversity of anomalies in each $\mathcal{D}_i$, $\mathcal{X}_{a,i}$ is randomly drawn from $\mathcal{X}_{a}$ and pseudo anomalies generated by popular anomaly generation methods \cite{li2021cutpaste,yun2019cutmix,zavrtanik2021draem}. 

Further, HADG utilizes those training data to create open-set detection and validation datasets for enabling the training of our model in a surrogate OSAD environment. Particularly, for each $\mathcal{D}_i$, HADG splits it into two disjoint subsets, \ie, $\mathcal{D}_i = \{\mathcal{D}^s_i, \mathcal{D}^q_i\}$, which correspond to support and query sets respectively, with the support set $\mathcal{D}_i^s=\mathcal{X}_{n,i}^s \cup \mathcal{X}_{a,i}^s$ used to train our base model $\phi_i$ and the query set $\mathcal{D}_i^q=\mathcal{X}_{n,i}^q \cup \mathcal{X}_{a,i}^q$ used to validate its open-set performance. To guarantee the openness in the validation/query set $\mathcal{D}_i^q$, we perform sampling in a way to ensure that $\mathcal{X}_{n,i}^s$ and $\mathcal{X}_{n,i}^q$ are two different normal clusters, and $\mathcal{X}_{a,i}^s$ and $\mathcal{X}_{a,i}^q$ do not overlap with each other, \ie, $\mathcal{X}_{a,i}^s \cap \mathcal{X}_{a,i}^q = \emptyset$.

\subsection{Collaborative Differentiable Learning of the Anomaly Heterogeneity}\label{subsec:cdl}

Our CDL component aims to first learn heterogeneous anomaly distributions hidden in $\mathcal{T}=\{\mathcal{D}_i\}_{i=1}^T$ by using $T$ base model $\phi_i$ and then utilize these models to collaboratively optimize the unified detection model $g$ in an end-to-end manner. CDL is presented in detail as follows.

\vspace{0.1cm}
\noindent\textbf{Learning $T$ Heterogeneous Anomaly Distributions.}

We first train $T$ base models $\{\phi_i\}_{i=1}^T$ to respectively capture heterogeneous anomaly distributions in $\{\mathcal{D}_i\}_{i=1}^T$, with each $\phi_i$ optimized using the following loss: 

\begin{equation}
    \mathcal{L}_{\phi_i} = \sum_{j=1}^{|\mathcal{D}^s_i|}\ell_{dev}\left(\phi_i(\mathbf{x}_j;\theta_i), y_j\right), 
\label{phi}
\end{equation}
where $\ell_{dev}$ is specified by a deviation loss~\cite{pang2021explainable}, following previous OSAD methods DRA~\cite{ding2022catching} and DevNet~\cite{pang2021explainable}, and $\mathcal{D}^s_i$ is the support set in $\mathcal{D}_i$. 
Although only limited seen anomalies are available during the training stage, the mixture of normal and anomaly samples in each $\mathcal{D}_i$ differs largely from each other, allowing each $\phi_i$ to learn a different anomaly distribution for anomaly scoring.

\vspace{0.1cm}
\noindent\textbf{Collaborative Differentiable Learning.}
Each $\phi_i$ captures only one part of the whole picture of the underlying anomaly heterogeneity. Thus, we then perform a collaborative differentiable learning that utilizes the losses from the $T$ base models to learn
the unified AD model $g$ for capturing richer anomaly heterogeneity. The key insight is that $g$ is optimized in a way to work well on a variety of possible anomaly distributions, mitigating potential overfitting to a particular anomaly distribution. Further, the optimization of $g$ is based on the losses on the query sets that are not seen during training the base models in Eq. \ref{phi}, \ie, $g$ is optimized under a surrogate open environment, which helps train a more generalized OSAD model $g$.

Specifically, $g$ is specified to have exactly the same network architecture as the based model $\phi_i$, and its weight parameters $\theta_{g}$ at epoch $t+1$ are optimized based on the losses resulted from all base models at epoch $t$:
\begin{equation}\label{eq:cdl}
\theta_{g}^{t} \longleftarrow \theta_{g}^{t-1} - \alpha \nabla\mathcal{L}_{\mathit{cdl}}, 
\end{equation}
where $\alpha$ is a learning rate and $\mathcal{L}_{\mathit{cdl}}$ is an aggregated loss over $T$ base models on the query sets:
\begin{equation}
\mathcal{L}_{\mathit{cdl}} = \sum_{i=1}^{T}\sum_{j=1}^{|\mathcal{D}^q_i|}\mathcal{L}_{\phi_i}\left(\phi_i(\mathbf{x}_j;\theta_i^t), y_j\right).
\label{heter optimization_initial}
\end{equation}
In the next training epoch, all $\theta_i^{t+1}$ of the base models are set to $\theta_g^{t}$ as their new weight parameters. We then optimize the base models $\phi_i$ using Eq. \ref{phi} on the support sets, and subsequently optimize the unified model $g$ using Eq. \ref{eq:cdl} on the query sets. This alternative base model and unified model learning is used to obtain an AD model $g$ that increasingly captures richer anomaly heterogeneity.

\vspace{0.1cm}
\noindent\textbf{Learning the Importance Scores of Individual Anomaly Distributions.}
The quality of the simulated anomaly distribution data $\mathcal{D}_i$ can vary largely, leading to base models of large difference in their effectiveness. 
Also, a base model that is less effective at one epoch can become more effective at another epoch.
Therefore, considering every base model equally throughout the optimization dynamic may lead to inferior optimization because poorly performing base models can affect the overall performance of the unified model $g$. 
To address this issue, we propose a self-supervised sequential modeling module to dynamically estimate the importance of each base model at each epoch. This refines the $\mathcal{L}_{\mathit{cdl}}$ loss as follows:
\begin{equation}
\mathcal{L}_{\mathit{cdl}}^{+} = \sum_{i=1}^{T}\sum_{j=1}^{|\mathcal{D}^q_i|}w_i^{t}\mathcal{L}_{\phi_i}\left(\phi_i(\mathbf{x}_j;\theta_i^t), y_j\right),
\label{heter optimization}
\end{equation}
where $w_i^t$ denotes the importance score of its base model $\phi_i$ at epoch $t$. Below we present how $w_i^t$ is learned via $\psi$.

Our sequential modeling-based estimation of the dynamic importance score is built upon the intuition that if a base model $\phi_i$ has good generalization ability, its predicted anomaly scores for different input data should be consistent and accurate at different training stages, where various anomaly heterogeneity gradually emerges as the training unfolds. To this end, wen train a sequential model $\psi$ to capture the consistency and accuracy of the anomaly scores yielded by all base models. This is achieved by training $\psi$ to predict the next epoch's anomaly scores of the base models using their previous output anomaly scores.
Specifically, given a training sample $\mathbf{x}_j$, the base models $\{\phi_i\}_{i=1}^{T}$ make a set of anomaly scoring predictions $\mathbf{s}_j =\{s_{ji}\} _{i=1}^T$,
resulting in a sequence of score predictions prior to an epoch $t$, $\mathbf{S}^\mathit{t}_j=[\mathbf{s}^{\mathit{t}-K}_j,\cdots,\mathbf{s}^{\mathit{t}-2}_j,\mathbf{s}^{t-1}_j]$ tracing back to $K$ previous steps, then $\psi:\mathbf{S}\rightarrow \mathbb{R}^{T}$ aims to predict the scoring predictions of all $T$ base models at epoch $t$. In our implementation, $\psi$ is specified by a sequential neural network parameterized by $\theta_\psi$, and it is optimized using the following next sequence prediction loss:

\begin{equation}\label{eq:sequential_model}
    \mathcal{L}_{\mathit{seq}} = \sum_{\mathbf{x}_j\in \mathcal{D}} \mathcal{L}_{\mathit{mse}}(\hat{\mathbf{s}}^{\mathit{t}}_j, \mathbf{s}^{\mathit{t}}_j),
\end{equation}
where $\hat{\mathbf{s}}^{\mathit{t}}_j=\psi(\mathbf{S}^t_j;\theta_\psi)$ and $\mathbf{s}^{\mathit{t}}_j$ are respectively the predicted and actual anomaly scores of $\mathbf{x}_j$ from the base models at epoch $t$, and $\mathcal{L}_{\mathit{seq}}$ is a mean squared error function. Instead of using supervised losses, the model $\psi$ is trained using a self-supervised loss function in Eq. \ref{eq:sequential_model} so as to withhold the ground truth labels for avoiding overfitting the labeled data and effectively evaluating the generalization ability of the base models. 

The generalization error $r_i^t$ for base model $\phi_i$ is then defined using the difference between the predicted anomaly score $\hat{s}^{t}_{ji}$ and the real label $y_j$ as follows:
\begin{equation}
     r_i^t = \frac{1}{|\mathcal{D}^\prime|}\sum_{\mathbf{x}_j\in \mathcal{D}^{\prime}} c_{j} \mathcal{L}_{\mathit{mse}}(\hat{s}_{ji}^{t}, y_j),
\label{r}
\end{equation}
where $\mathcal{D}^{\prime}= \mathcal{D}\setminus \mathcal{X}_{n,i}$ and $c_{j}$ is a pre-defined category-wise weight associate with each example $\mathbf{x}_j$. In other words, $r_i^t$ measures the detection error of $\phi_i$ in predicting the anomaly scores for the seen anomalies in $\mathcal{X}_{a,i}$ and all other unseen normal and anomaly training examples, excluding the seen normal examples $\mathcal{X}_{n,i}$ w.r.t. $\phi_i$. A larger $c_{j}$ is assigned if $\mathbf{x}_j$ is an unseen anomaly to highlight the importance of detecting unseen anomalies; and it is assigned with the same value for the other examples otherwise.

Since a large $r_i^t$ implies a poor generalization ability of the base model $\phi_i$ at epoch $t$, we should pay less attention to it when updating the unified model $g$. Therefore, the importance score of $\phi_i$ is defined as the inverse of its generalization error as follows:
\begin{equation}
     w_i^t = \frac{\exp(-r_i^t)}{\sum_i^T \exp(-r_i^t)}.
\label{w}
\end{equation}

\section{Experiments}
\label{sec:experiment}

\begin{table*}[ht]
\centering
\resizebox{0.9\textwidth}{!}{
\begin{tabular}{|cccccc|cc|}
\hline
\multicolumn{1}{|c|}{\textbf{Dataset}}    & \textbf{SAOE} & \textbf{MLEP} & \textbf{FLOS} & \textbf{DevNet}      & \textbf{DRA}         & \textbf{\coolname(DevNet)}                      & \multicolumn{1}{c|}{\textbf{\coolname(DRA)}}                        \\ \hline
\multicolumn{8}{|c|}{\textbf{Ten Anomaly Examples (Random)}}                                                                                                                                                                                                 \\ \hline
\multicolumn{1}{|c|}{\textbf{AITEX}}         & 0.874{\scriptsize±0.024}   & 0.867{\scriptsize±0.037}   & 0.841{\scriptsize±0.049}   & 0.889{\scriptsize±0.007}          & 0.892{\scriptsize±0.007}          & \textbf{0.903{\scriptsize±0.011}}$\textcolor{magenta}\uparrow$                        & \multicolumn{1}{c|}{\color[HTML]{FF0000} \textbf{0.925{\scriptsize±0.013}}$\textcolor{blue}\uparrow$} \\
\multicolumn{1}{|c|}{\textbf{SDD}}           & 0.955{\scriptsize±0.020}   & 0.783{\scriptsize±0.013}   & 0.967{\scriptsize±0.018}   & 0.985{\scriptsize±0.004}          & \textbf{0.990{\scriptsize±0.000}} & {\color[HTML]{FF0000} \textbf{0.991{\scriptsize±0.001}}}$\textcolor{magenta}\uparrow$  & \multicolumn{1}{c|}{\color[HTML]{FF0000} \textbf{0.991{\scriptsize±0.000}}$\textcolor{blue}\uparrow$} \\
\multicolumn{1}{|c|}{\textbf{ELPV}}          & 0.793{\scriptsize±0.047}   & 0.794{\scriptsize±0.047}   & 0.818{\scriptsize±0.032}   & 0.843{\scriptsize±0.001}          & 0.843{\scriptsize±0.002}          & \textbf{0.849{\scriptsize±0.003}}$\textcolor{magenta}\uparrow$                        & \multicolumn{1}{c|}{\color[HTML]{FF0000} \textbf{0.850{\scriptsize±0.004}}$\textcolor{blue}\uparrow$} \\
\multicolumn{1}{|c|}{\textbf{Optical}}       & 0.941{\scriptsize±0.013}   & 0.740{\scriptsize±0.039}   & 0.720{\scriptsize±0.055}   & 0.785{\scriptsize±0.012}          & \textbf{0.966{\scriptsize±0.002}} & 0.841{\scriptsize±0.010}$\textcolor{magenta}\uparrow$                                  & \multicolumn{1}{c|}{\color[HTML]{FF0000} \textbf{0.976{\scriptsize±0.004}}$\textcolor{blue}\uparrow$} \\
\multicolumn{1}{|c|}{\textbf{Mastcam}}       & 0.810{\scriptsize±0.029}   & 0.798{\scriptsize±0.026}   & 0.703{\scriptsize±0.029}   & 0.797{\scriptsize±0.021}          & \textbf{0.849{\scriptsize±0.003}} & 0.825{\scriptsize±0.020}$\textcolor{magenta}\uparrow$                                  & \multicolumn{1}{c|}{\color[HTML]{FF0000} \textbf{0.855{\scriptsize±0.005}}$\textcolor{blue}\uparrow$} \\
\multicolumn{1}{|c|}{\textbf{BrainMRI}}      & 0.900{\scriptsize±0.041}   & 0.959{\scriptsize±0.011}   & 0.955{\scriptsize±0.011}   & 0.951{\scriptsize±0.007}          & \textbf{0.971{\scriptsize±0.001}} & 0.959{\scriptsize±0.008}$\textcolor{magenta}\uparrow$                                  & \multicolumn{1}{c|}{\color[HTML]{FF0000} \textbf{0.977{\scriptsize±0.001}}$\textcolor{blue}\uparrow$} \\
\multicolumn{1}{|c|}{\textbf{HeadCT}}        & 0.935{\scriptsize±0.021}   & 0.972{\scriptsize±0.014}   & 0.971{\scriptsize±0.004}   & 0.997{\scriptsize±0.002}          & 0.978{\scriptsize±0.001}          & {\color[HTML]{FF0000} \textbf{0.999{\scriptsize±0.003}}}$\textcolor{magenta}\uparrow$  & \multicolumn{1}{c|}{\textbf{0.993{\scriptsize±0.002}}$\textcolor{blue}\uparrow$}                        \\
\multicolumn{1}{|c|}{\textbf{Hyper-Kvasir}}  & 0.666{\scriptsize±0.050}   & 0.600{\scriptsize±0.069}   & 0.773{\scriptsize±0.029}   & 0.822{\scriptsize±0.031}          & 0.844{\scriptsize±0.001}          & \textbf{0.873{\scriptsize±0.009}}$\textcolor{magenta}\uparrow$                         & \multicolumn{1}{c|}{\color[HTML]{FF0000} \textbf{0.880{\scriptsize±0.003}}$\textcolor{blue}\uparrow$} \\ 
\multicolumn{1}{|c|}{\textbf{MVTec AD {\color[HTML]{808080} {\scriptsize(mean)}}}}  & 0.926{\scriptsize±0.010}   & 0.907{\scriptsize±0.005}   & 0.939{\scriptsize±0.007}   & 0.948{\scriptsize±0.005}          & \textbf{0.966{\scriptsize±0.002}} & 0.954{\scriptsize±0.003}$\textcolor{magenta}\uparrow$                                  & \multicolumn{1}{c|}{\color[HTML]{FF0000} \textbf{0.970{\scriptsize±0.002}}$\textcolor{blue}\uparrow$} \\ \hline
\multicolumn{8}{|c|}{\textbf{One Anomaly Example (Random)}}                                                                                                                                                                                                  \\ \hline
\multicolumn{1}{|c|}{\textbf{AITEX}}         & 0.675{\scriptsize±0.094}   & 0.564{\scriptsize±0.055}   & 0.538{\scriptsize±0.073}   & 0.609{\scriptsize±0.054}          & 0.693{\scriptsize±0.031}          & \textbf{0.704{\scriptsize±0.004}}$\textcolor{magenta}\uparrow$                         & \multicolumn{1}{c|}{\color[HTML]{FF0000} \textbf{0.734{\scriptsize±0.008}}$\textcolor{blue}\uparrow$} \\
\multicolumn{1}{|c|}{\textbf{SDD}}           & 0.781{\scriptsize±0.009}   & 0.811{\scriptsize±0.045}   & 0.840{\scriptsize±0.043}   & 0.851{\scriptsize±0.003}          & \textbf{0.907{\scriptsize±0.002}} & 0.864{\scriptsize±0.001}$\textcolor{magenta}\uparrow$                                  & \multicolumn{1}{c|}{\color[HTML]{FF0000} \textbf{0.909{\scriptsize±0.001}}$\textcolor{blue}\uparrow$} \\
\multicolumn{1}{|c|}{\textbf{ELPV}}          & 0.635{\scriptsize±0.092}   & 0.578{\scriptsize±0.062}   & 0.457{\scriptsize±0.056}   & \textbf{0.810{\scriptsize±0.024}} & 0.676{\scriptsize±0.003}          & \multicolumn{1}{c}{\color[HTML]{FF0000}\textbf {0.828{\scriptsize±0.005}}$\textcolor{magenta}\uparrow$ } & \multicolumn{1}{c|}{0.723{\scriptsize±0.008}$\textcolor{blue}\uparrow$}                          \\
\multicolumn{1}{|c|}{\textbf{Optical}}      & 0.815{\scriptsize±0.014}   & 0.516{\scriptsize±0.009}   & 0.518{\scriptsize±0.003}   & 0.513{\scriptsize±0.001}          & \textbf{0.880{\scriptsize±0.002}} & 0.547{\scriptsize±0.009}$\textcolor{magenta}\uparrow$                                  & \multicolumn{1}{c|}{\color[HTML]{FF0000} \textbf{0.888{\scriptsize±0.007}}$\textcolor{blue}\uparrow$} \\
\multicolumn{1}{|c|}{\textbf{Mastcam}}       & 0.662{\scriptsize±0.018}   & 0.625{\scriptsize±0.045}   & 0.542{\scriptsize±0.017}   & 0.627{\scriptsize±0.049}          & \textbf{0.709{\scriptsize±0.011}} & 0.644{\scriptsize±0.013}$\textcolor{magenta}\uparrow$               & \multicolumn{1}{c|}{\color[HTML]{FF0000} \textbf{0.743{\scriptsize±0.003}}$\textcolor{blue}\uparrow$} \\
\multicolumn{1}{|c|}{\textbf{BrainMRI}}      & 0.531{\scriptsize±0.060}   & 0.632{\scriptsize±0.017}   & 0.693{\scriptsize±0.036}   & \textbf{0.853{\scriptsize±0.045}} & 0.747{\scriptsize±0.001}          & 
{\color[HTML]{FF0000} \textbf{0.866{\scriptsize±0.004}}}$\textcolor{magenta}\uparrow$  & \multicolumn{1}{c|}{0.760{\scriptsize±0.013}$\textcolor{blue}\uparrow$}                                \\
\multicolumn{1}{|c|}{\textbf{HeadCT}}        & 0.597{\scriptsize±0.022}   & 0.758{\scriptsize±0.038}   & 0.698{\scriptsize±0.092}   & 0.755{\scriptsize±0.029}          & \textbf{0.804{\scriptsize±0.010}} & 0.781{\scriptsize±0.007}$\textcolor{magenta}\uparrow$                  & \multicolumn{1}{c|}{\color[HTML]{FF0000} \textbf{0.825{\scriptsize±0.014}}$\textcolor{blue}\uparrow$} \\
\multicolumn{1}{|c|}{\textbf{Hyper-Kvasir}}  & 0.498{\scriptsize±0.100}   & 0.445{\scriptsize±0.040}   & 0.668{\scriptsize±0.004}   & 0.734{\scriptsize±0.020}          & 0.712{\scriptsize±0.010}          & {\color[HTML]{FF0000} \textbf{0.768{\scriptsize±0.015}}}$\textcolor{magenta}\uparrow$  & \multicolumn{1}{c|}{\textbf{0.742{\scriptsize±0.015}}$\textcolor{blue}\uparrow$}                        \\
\multicolumn{1}{|c|}{\textbf{MVTec AD {\color[HTML]{808080} {\scriptsize(mean)}}}}  & 0.834{\scriptsize±0.007}   & 0.744{\scriptsize±0.019}   & 0.792{\scriptsize±0.014}   & 0.832{\scriptsize±0.016}          & \textbf{0.889{\scriptsize±0.013}} & 0.843{\scriptsize±0.021}$\textcolor{magenta}\uparrow$                                  & \multicolumn{1}{c|}{\color[HTML]{FF0000} \textbf{0.901{\scriptsize±0.003}}$\textcolor{blue}\uparrow$} \\
\hline
\end{tabular}}
\caption{AUC results(mean±std) on nine real-world AD datasets under the general setting. Best results and the second-best results are respectively highlighted in {\color[HTML]{FF0000} \textbf{red}} and \textbf{bold}.`$\textcolor{blue}\uparrow$' (`$\textcolor{magenta}\uparrow$') indicates increased performance over DRA (DevNet).}
\label{general_result}
\vspace{-0.5cm}
\end{table*}

\subsection{Experimental Setup}

\vspace{0.1cm}
\noindent\textbf{Datasets.} 
Following prior OSAD studies \cite{pang2021explainable,ding2022catching}, we conduct extensive experiments on nine real-world anomaly detection datasets, including five industrial defect inspection datasets (MVTec AD~\cite{bergmann2019mvtec}, AITEX~\cite{silvestre2019public}, SDD~\cite{tabernik2020segmentation}, ELPV~\cite{deitsch2019automatic} and Optical~\cite{wieler2007weakly}), one planetary exploration dataset (Mastcam~\cite{kerner2020comparison}), and three medical datasets (HeadCT~\cite{salehi2021multiresolution}, BrainMRI~\cite{salehi2021multiresolution} and Hyper-Kvasir~\cite{borgli2020hyperkvasir}). Depending on how we sample the seen anomaly examples, two protocols are used to evaluate the detection performance, the general and hard settings~\cite{ding2022catching}. The \textit{general setting} assumes the anomaly examples are randomly sampled from the anomaly classes, while the \textit{hard setting} presents a more challenging case where the anomaly examples are sampled exclusively from only one class to evaluate the generalization ability to novel or unseen anomaly classes. Both protocols are used in our experiments. Following \cite{ding2022catching}, we also evaluate the performance with the number of anomaly examples set to respectively $M=10$ and $M=1$.
Further details about these datasets are available in \texttt{Appendix A}.

\vspace{0.1cm}
\noindent\textbf{Competing Methods and Evaluation Metrics.}
\coolname is compared with five closely related state-of-the-art (SOTA) methods, including MLEP~\cite{liu2019margin}, SAOE~\cite{markovitz2020graph, li2021cutpaste, tack2020csi}, FLOS~\cite{lin2017focal}, DevNet \cite{pang2021explainable}, and DRA \cite{ding2022catching}. MLEP, DevNet and DRA are specifically designed for OSAD. SAOE is a supervised detector augmented with synthetic anomalies and outlier exposure, while FLOS is a focal-loss-based imbalanced classifier. 
For evaluation metrics, we adopt the widely used Area Under ROC Curve (AUC) to measure the performance of all methods and settings. All reported results are averaged results over three independent runs, and stated otherwise.

\vspace{0.1cm}
\noindent\textbf{Implementation Details.}
To generate a diverse set of anomaly distributions, our proposed approach uses a mixture of randomly selected normal clusters and labeled anomaly examples to create each individual anomaly distribution data $\mathcal{D}_i$. Specifically, $k$-means clustering is first used to partition normal samples into three normal clusters (\ie, $k=3$ is used). Then two randomly selected clusters are chosen, combining with the seen anomalies, to construct $\mathcal{D}_i$, having one normal clusters and $50\%$ of the seen anomaly set as the support set $\mathcal{D}_i^s$ while the rest of samples are used as the query set $\mathcal{D}_i^q$ (Under the protocol of having only one seen anomaly example, the example is included in both sets). This helps effectively simulate open-set environments with partially observed anomaly distributions. To further increase the heterogeneity in within and between the anomaly distribution datasets, we randomly pick one of the three popular anomaly generation techniques, including CutMix~\cite{yun2019cutmix}, CutPaste~\cite{li2021cutpaste}, and DRAEM Mask~\cite{zavrtanik2021draem}, to generate and inject pseudo anomalies into the support and query sets of $\mathcal{D}_i$. To guarantee the openness w.r.t. the pseudo anomaly detection, the pseudo anomalies in $\mathcal{D}_i^s$ and $\mathcal{D}_i^q$ are generated from two different anomaly generation approaches. For each dataset, $T=7$ is used in generating the individual anomaly distribution data.
$c_{j}$ is set to 1.0 when $\mathbf{x}_j$ represents unseen anomaly samples, and 0.5 when $\mathbf{x}_j$ represents seen anomaly or unseen normal samples.

\coolname is a generic framework, under which features and loss functions from existing OSAD models can be easily plugged in as the base features and the base loss. 
Particularly, the image features are extracted from one of the OSAD model (\eg, DRA), and then \coolname is trained using 
our proposed loss function built on the base loss (see Eq. \ref{heter optimization}). DRA~\cite{ding2022catching}, DevNet~\cite{pang2021explainable} and BGAD \cite{yao2023explicit} are the current SOTA models for OSAD, but BGAD uses quite different benchmark datasets from the other two. Our experiments strictly follow the seminal OSAD evaluation protocol and benchmarks used in DRA~\cite{ding2022catching} and DevNet~\cite{pang2021explainable}, and choose DRA~\cite{ding2022catching} and DevNet~\cite{pang2021explainable} to respectively plug in \coolname, denoting as \textbf{\coolname (DRA)} and \textbf{\coolname (DevNet)}.
Adam is used as the optimizer. The initial learning rate for learning heterogeneous $T$ base models is set to 0.0002, while that for the unified AD model $g$ is set to 0.005.
In the self-supervised importance score estimator, 
a two-layer Bidirectional LSTM~\cite{zhou2016attention} is used as the backbone, with the hidden dimension set to 7. 
It is followed by a fully-connected layer with 14 hidden nodes, before the prediction layer. 
The initial learning rate is set to 0.02 for this component.

The above settings are used by default for the reported results of \coolname across all the datasets. The results of MLEP, SAOE, and FLOS are taken from \cite{ding2022catching}. The results of DevNet and DRA are reproduced using their official codes to obtain the features used in \coolname, meaning that DevNet and \coolname (DevNet) use the same set of features, which also applies to DRA and \coolname (DRA) (see \texttt{Appendix B} for more implementation details). 

\begin{table*}[ht]
\centering
\resizebox{0.9\textwidth}{!}{
\begin{tabular}{|cccccc|cc|}
\hline
\multicolumn{1}{|c|}{\textbf{Dataset}}   & \multicolumn{1}{c}{\textbf{SAOE}} & \multicolumn{1}{c}{\textbf{MLEP}} & \multicolumn{1}{c}{\textbf{FLOS}} & \textbf{DevNet} & \textbf{DRA}                                 & \textbf{\coolname(DevNet)}                       & \textbf{\coolname(DRA)}                           \\ \hline
\multicolumn{8}{|c|}{\textbf{Ten Examples from One Anomaly Class}}                                                                                                                                                                                                                                                                                                                                                                                                                 \\ \hline
\multicolumn{1}{|c|}{\textbf{Carpet {\color[HTML]{808080} {\scriptsize(mean)}}}}                  & 0.762{\scriptsize±0.073}                                               & 0.935{\scriptsize±0.013}                                               & 0.761{\scriptsize±0.012}                                               & 0.853{\scriptsize±0.005}     & \textbf{0.940{\scriptsize±0.006}}                         & 0.860{\scriptsize±0.013}$\textcolor{magenta}\uparrow$                                  & {\color[HTML]{FF0000} \textbf{0.949{\scriptsize±0.002}}$\textcolor{blue}\uparrow$}   \\
\multicolumn{1}{|c|}{\textbf{Metal\_nut {\color[HTML]{808080} {\scriptsize(mean)}}}}                & 0.855{\scriptsize±0.016}                                               & 0.945{\scriptsize±0.017}                                               & 0.922{\scriptsize±0.014}                                               & 0.970{\scriptsize±0.009}     & 0.968{\scriptsize±0.006}                                  & {\color[HTML]{FF0000} \textbf{0.972{\scriptsize±0.002}}$\textcolor{magenta}\uparrow$}  & \textbf{0.971{\scriptsize±0.001}$\textcolor{blue}\uparrow$}                          \\
\multicolumn{1}{|c|}{\textbf{AITEX {\color[HTML]{808080} {\scriptsize(mean)}}}}                  & 0.724{\scriptsize±0.032}                                               & 0.733{\scriptsize±0.009}                                               & 0.635{\scriptsize±0.043}                                               & 0.685{\scriptsize±0.016}     & \textbf{0.733{\scriptsize±0.011}}                         & 0.709{\scriptsize±0.006}$\textcolor{magenta}\uparrow$                                  & {\color[HTML]{FF0000} \textbf{0.747{\scriptsize±0.002}}$\textcolor{blue}\uparrow$}   \\
\multicolumn{1}{|c|}{\textbf{ELPV {\color[HTML]{808080} {\scriptsize(mean)}}}}                 & 0.683{\scriptsize±0.047}                                               & 0.766{\scriptsize±0.029}                                               & 0.646{\scriptsize±0.032}                                               & 0.722{\scriptsize±0.018}     & \textbf{0.771{\scriptsize±0.005}}                         & 0.752{\scriptsize±0.005}$\textcolor{magenta}\uparrow$                                  & {\color[HTML]{FF0000} \textbf{0.788{\scriptsize±0.003}}$\textcolor{blue}\uparrow$}   \\
\multicolumn{1}{|c|}{\textbf{Mastcam {\color[HTML]{808080} {\scriptsize(mean)}}}}              & 0.697{\scriptsize±0.014}                                               & 0.695{\scriptsize±0.004}                                               & 0.616{\scriptsize±0.021}                                               & 0.588{\scriptsize±0.025}     & \textbf{0.704{\scriptsize±0.007}}                         & 0.602{\scriptsize±0.008}$\textcolor{magenta}\uparrow$                                  & {\color[HTML]{FF0000} \textbf{0.721{\scriptsize±0.003}}$\textcolor{blue}\uparrow$}   \\
\multicolumn{1}{|c|}{\textbf{Hyper-Kvasir {\color[HTML]{808080} {\scriptsize(mean)}}}}             & 0.698{\scriptsize±0.021}                                               & 0.844{\scriptsize±0.009}                                               & 0.786{\scriptsize±0.021}                                               & 0.827{\scriptsize±0.008}     & 0.822{\scriptsize±0.013}                                  & \textbf{0.845{\scriptsize±0.003}}$\textcolor{magenta}\uparrow$                         & {\color[HTML]{FF0000} \textbf{0.854{\scriptsize±0.004}}$\textcolor{blue}\uparrow$}   \\ \hline
\multicolumn{8}{|c|}{\textbf{One Example from One Anomaly Class}}                                                                                                                                                                                                                                                                                                                                                                                                                  \\ \hline
\multicolumn{1}{|c|}{\textbf{Carpet {\color[HTML]{808080} {\scriptsize(mean)}}}}                        & 0.753{\scriptsize±0.055 }                                              & 0.679{\scriptsize±0.029 }                                              & 0.678{\scriptsize±0.040}                                               & 0.774{\scriptsize±0.007}     & \textbf{0.905{\scriptsize±0.006}}                         & 0.785{\scriptsize±0.015}$\textcolor{magenta}\uparrow$                                  & {\color[HTML]{FF0000} \textbf{0.932{\scriptsize±0.003}}$\textcolor{blue}\uparrow$}   \\
\multicolumn{1}{|c|}{\textbf{Metal\_nut {\color[HTML]{808080} {\scriptsize(mean)}}}}               & 0.816{\scriptsize±0.029}                                               & 0.825{\scriptsize±0.023}                                               & 0.855{\scriptsize±0.024 }                                              & 0.861{\scriptsize±0.019}     & \textbf{0.936{\scriptsize±0.011}}                         & 0.869{\scriptsize±0.004}$\textcolor{magenta}\uparrow$                                  & {\color[HTML]{FF0000} \textbf{0.939{\scriptsize±0.004}}$\textcolor{blue}\uparrow$}   \\
\multicolumn{1}{|c|}{\textbf{AITEX {\color[HTML]{808080} {\scriptsize(mean)}}}}                   & 0.674{\scriptsize±0.034 }                                              & 0.466{\scriptsize±0.030}                                               & 0.624{\scriptsize±0.024}                                               & 0.646{\scriptsize±0.014}     & \textbf{0.696{\scriptsize±0.011}}                                  & 0.660{\scriptsize±0.007}$\textcolor{magenta}\uparrow$                                  & \multicolumn{1}{c|}{\color[HTML]{FF0000} \textbf{0.707{\scriptsize±0.007}}$\textcolor{blue}\uparrow$} \\
\multicolumn{1}{|c|}{\textbf{ELPV {\color[HTML]{808080} {\scriptsize(mean)}}}}                   & 0.614{\scriptsize±0.048}                                               & 0.566{\scriptsize±0.111}                                               & 0.691{\scriptsize±0.008}                                               & 0.663{\scriptsize±0.008}     & \multicolumn{1}{c|}{\textbf{0.722{\scriptsize±0.006}}} & 0.678{\scriptsize±0.006}$\textcolor{magenta}\uparrow$                                  & \multicolumn{1}{c|}{\color[HTML]{FF0000} \textbf{0.740{\scriptsize±0.003}}$\textcolor{blue}\uparrow$} \\
\multicolumn{1}{|c|}{\textbf{Mastcam {\color[HTML]{808080} {\scriptsize(mean)}}}}                & 0.689{\scriptsize±0.037}                                               & 0.541{\scriptsize±0.007}                                               & 0.524{\scriptsize±0.013}                                               & 0.519{\scriptsize±0.014}     & \multicolumn{1}{c|}{\textbf{0.658{\scriptsize±0.021}}} & \multicolumn{1}{c}{0.535{\scriptsize±0.003}$\textcolor{magenta}\uparrow$} & \multicolumn{1}{c|}{\color[HTML]{FF0000} \textbf{0.673{\scriptsize±0.010}}$\textcolor{blue}\uparrow$} \\
\multicolumn{1}{|c|}{\textbf{Hyper-Kvasir {\color[HTML]{808080} {\scriptsize(mean)}}}}         & 0.406{\scriptsize±0.018}                                               & 0.480{\scriptsize±0.044}                                               & 0.571{\scriptsize±0.004}                                               & 0.598{\scriptsize±0.006}     & \multicolumn{1}{c|}{\textbf{0.699{\scriptsize±0.009}}} & 0.619{\scriptsize±0.005}$\textcolor{magenta}\uparrow$                                  & \multicolumn{1}{c|}{\color[HTML]{FF0000} \textbf{0.706{\scriptsize±0.007}}$\textcolor{blue}\uparrow$} \\ \hline
\end{tabular}}
\caption{AUC results(mean±std) on nine real-world AD datasets under the hard setting. Best results and the second-best results are respectively highlighted in {\color[HTML]{FF0000} \textbf{red}} and \textbf{bold}.`$\textcolor{blue}\uparrow$' (`$\textcolor{magenta}\uparrow$') indicates increased performance over DRA (DevNet). Carpet and Meta\_nut are two subsets of MVTec AD. The same set of datasets is used as in \cite{ding2022catching}.}
\label{hard_result}
\vspace{-0.5cm}
\end{table*}

\subsection{Performance under General Settings}
Table~\ref{general_result} shows the comparison results under the general setting, where models are trained using one or ten randomly sampled anomaly examples. The results on MVTec AD are averaged over its 16 data subsets (see \texttt{Appendix C} for detailed results on the subsets).
Overall, our approach \coolname brings consistently substantial improvement to the respective DRA and DevNet in both ten-shot and one-shot setting protocols across all the datasets of three application scenarios. Since DRA is a stronger base model than DevNet, \coolname (DRA) generally obtains much better performance than \coolname (DevNet).
On average, \coolname respectively enhances the performance of DRA and DevNet by up to $4\%$ and $9\%$ AUC,
indicating that the anomaly heterogeneity learned by \coolname enable the base models to gain significantly improved generalization ability.

\subsection{Generalization to Unseen Anomaly Classes}
Table~\ref{hard_result} shows the comparison results under the hard setting, where models are trained with one or ten examples randomly sampled from exclusively one known anomaly class to detect the anomalies in the rest of all other anomaly classes. This protocol means that we can obtain one AUC result for having each anomaly class as the seen anomaly class. The reported results here are averaged over all the anomaly classes  (see \texttt{Appendix C} for detailed class-level results). 
In general, our models -- \coolname (DRA) and \coolname (DevNet) -- achieves the best AUC results in both $M=1$ and $M=10$ settings. To be more precise, \coolname respectively improves the performances of DRA and DevNet by up to $3.2\%$ and $3\%$ AUC. Similar to the general setting, \coolname (DRA) is the best performer. Due to the fact that the model is only exposed to one single anomaly class, the improvement is fully due to the ability of \coolname in generalizing to detect unseen anomaly classes. 

\subsection{Unseen Anomaly Detection in Novel Domains}
We evaluate the effectiveness of \coolname in generalizing to unseen anomalies in a novel domain (\ie, a cross-domain AD task), where a model is trained on a source domain and is subsequently tested on datasets from a target domain that differs from the source. This is used to further verify the generalization capacity of \coolname. Specifically, following the DRA work \cite{ding2022catching}, we employ the DRA and \coolname(DRA) models, trained on one of five datasets (the source domain) and fine-tune them for 10 epochs using only normal samples on the other four datasets (the target domains). The results of this experiment are presented in Table~\ref{cross}. Overall, the \coolname(DRA) model outperforms the DRA baseline significantly in this setting, and it can even achieves comparable AUC to the same-domain performance (\ie, the diagonal results). However, we do observe a slight drop in performance on some target domains. We attribute this to the learned anomaly heterogeneity being based on anomaly samples from the source domain, which may introduce bias when testing the AD model on the target domains. Nevertheless, our findings indicate that the \coolname framework enhances the generalization ability of DRA on novel domains, demonstrating promising cross-domain performance.

\begin{table}[t]
\resizebox{\linewidth}{!}{
\begin{tabular}{|c|ccccc|ccccc|}
\hline
\multirow{2}{*}{} & \multicolumn{5}{c|}{\textbf{DRA}}                 & \multicolumn{5}{c|}{\textbf{\coolname(DRA)}}         \\ \cline{2-11} 
                  & \textbf{Carpet} & \textbf{Grid}  & \textbf{Leather} & \textbf{Tile}  & \textbf{Wood} & \textbf{Carpet} & \textbf{Grid} & \textbf{Leather} & \textbf{Tile} & \textbf{Wood} \\ \hline
\textbf{Carpet}            & \color[HTML]{A9A9A9}{0.945} & 0.833 & 0.921   & 0.930 & 0.917 & \color[HTML]{A9A9A9}{0.953} & \textbf{0.979}  & \textbf{1.000} & \textbf{1.000} & \textbf{0.998}\\
\textbf{Grid}              & \textbf{0.983}  & \color[HTML]{A9A9A9}{0.990}     & 0.924   & 0.940 & 0.916 & 0.959  & \color[HTML]{A9A9A9}{0.992}      & \textbf{1.000}   & \textbf{1.000} & \textbf{0.973}\\
\textbf{Leather}           & \textbf{0.988}  & 0.998 & \color[HTML]{A9A9A9}{1.000}       & 0.994 & \textbf{1.000} & 0.963  & \textbf{0.998}  & \color[HTML]{A9A9A9}{1.000}       & \textbf{1.000} & 0.998\\
\textbf{Tile}              & 0.917  & 0.971 & 0.958   & \color[HTML]{A9A9A9}{1.000}     & 0.955 & \textbf{0.943}  & \textbf{0.982}  & \textbf{1.000}   & \color[HTML]{A9A9A9}{1.000}     & \textbf{0.999}\\
\textbf{Wood}              & 0.993  & 0.985 & 0.972   & 0.948 & \color[HTML]{A9A9A9}{0.998}     & \textbf{0.995}  & \textbf{0.989}  & \textbf{1.000}   & \textbf{1.000} & \color[HTML]{A9A9A9}{0.998}    \\ \hline
\end{tabular}}
\caption{AUC results of DRA and \coolname(DRA) on detecting texture anomalies in cross-domain datasets (all are MVTec AD datasets). The top row is the source domain for training, while the left column is the target domain  for inference. The same-domain results are shown in $\color[HTML]{A9A9A9}{gray}$ for reference. Best results are \textbf{boldfaced}.}
\label{cross}
\vspace{-0.5cm}
\end{table}

\subsection{Analysis of \textbf{\coolname}}

\subsubsection{Utility of Few-shot Samples}
To investigate the utility of few-shot samples in the OSAD task, \coolname is also compared with state-of-the-art models of unsupervised anomaly detection (UAD) and fully-supervised anomaly detection (FSAD),
including KDAD~\cite{salehi2021multiresolution} and PatchCore~\cite{roth2022towards} for UAD and fully supervised DRA (FS-DRA for short) for FSAD. DRA is transformed to a fully supervised approach FS-DRA by using a set of 10 anomaly examples that illustrate all possible anomaly classes in the test data. The results of \coolname (DRA) using randomly sampled 10 anomaly examples are used here for comparison. 
The results are presented in Table~\ref{Utility}. It shows that \coolname (DRA) substantially outperforms the unsupervised detectors KDAD and PatchCore, demonstrating better generalization ability than the unsupervised methods. FS-DRA is the best-performing model on six out of nine datasets. This is expected due to its fully supervised nature. 
Although \coolname (DRA) is an open-set detector, it is the best performer on AITEX and SDD and performs comparably well to the fully-supervised model FS-DRA on the other seven datasets, indicating that \coolname (DRA) can accurately generalize to detect unseen anomalies while effectively maintaining the effectiveness on the seen anomalies. These results suggest very effective utilization of the few-shot anomaly examples in \coolname, while avoiding the overfitting of the seen anomalies.

\begin{table}[t]
\centering
\resizebox{\linewidth}{!}{
\begin{tabular}{|c|cccc|}
\hline
\textbf{Dataset}           & \textbf{KDAD}    & \textbf{PatchCore} & \textbf{\coolname(DRA)}        & \textbf{FS-DRA} \\ \hline
\textbf{AITEX}             & 0.576{\scriptsize±0.002}         & 0.783     & {\color[HTML]{FF0000} \textbf{0.925{\scriptsize±0.013}}} & \multicolumn{1}{c|} {\textbf{0.919{\scriptsize±0.004}}}           \\
\textbf{SDD}               & 0.842{\scriptsize±0.006}          & \textbf{0.873}              & {\color[HTML]{FF0000} \textbf{0.991{\scriptsize±0.000}}} &    {\color[HTML]{FF0000}\textbf{0.991{\scriptsize±0.000}}}                                  \\
\textbf{ELPV}     & 0.744{\scriptsize±0.001}          & {\color[HTML]{FF0000}\textbf{0.916}}              & 0.850{\scriptsize±0.004} & \multicolumn{1}{c|}{\textbf{0.874{\scriptsize±0.004}}}                     \\
\textbf{Optical}  & 0.579{\scriptsize±0.002} & -                  & \textbf{0.976{\scriptsize±0.004}} & \multicolumn{1}{c|}{\color[HTML]{FF0000}{\textbf{0.982{\scriptsize±0.000}}}}                     \\
\textbf{Mastcam}  & 0.642{\scriptsize±0.007} & 0.809              &  \textbf{0.855{\scriptsize±0.005}} &  {\color[HTML]{FF0000}\textbf{0.877{\scriptsize±0.003 }}}                                                                \\
\textbf{BrainMRI} & 0.733{\scriptsize±0.016} & 0.754              & \textbf{0.977{\scriptsize±0.001}} &   {\color[HTML]{FF0000}\textbf{0.979{\scriptsize±0.001}}}                                              \\
\textbf{HeadCT}            & 0.793{\scriptsize±0.017}         & 0.864     &  \textbf{0.993{\scriptsize±0.002}} & \multicolumn{1}{c|}{\color[HTML]{FF0000}{\textbf{0.998{\scriptsize±0.003}}}}            \\
\textbf{Hyper-Kvasir}      & 0.401{\scriptsize±0.002}        & 0.494              & \textbf{0.880{\scriptsize±0.003}} & {\color[HTML]{FF0000}\textbf{0.900{\scriptsize±0.009}}}     \\ 
\textbf{MVTec AD {\color[HTML]{808080} {\scriptsize(mean)}}}      & 0.863{\scriptsize±0.029}          & {\color[HTML]{FF0000}\textbf{0.992}}     & 0.970{\scriptsize±0.002} & \multicolumn{1}{c|}{\textbf{0.984{\scriptsize±0.004}}}                       \\
\hline
\end{tabular}}
\caption{AUC results comparison of \coolname (DRA) to unsupervised anomaly detection methods -- KDAD and PatchCore -- and a fully-supervised DRA model (FS-DRA). Best results and the second-best results are respectively highlighted in {\color[HTML]{FF0000} \textbf{red}} and \textbf{bold}}
\label{Utility}
\vspace{-0.5cm}
\end{table}

\subsubsection{Ablation Study}\label{subsec:ablation}
Our ablation study uses DRA as the baseline. To evaluate our first component HADG, we compare DRA using only HADG (\textbf{+ HADG}) with its three variants, including DRA using random heterogeneous anomaly distribution generation (\textbf{+ RamHADG}), \ie, an ensemble of DRA trained on different random data subsets; an ensemble of DRA trained on the full data using different random seeds (\textbf{+ RamFULL}); an ensemble of heterogeneous detectors DRA and DevNet trained using RamHADG (\textbf{+ RamHADG + DevNet}). We then examine the effectiveness of another component CDL by adding it on top of the variant DRA + HADG, which is also our full \coolname model, denoted as `\textbf{+ CDL}' for brevity. `\textbf{+ CDL}' is also compared with its simplified version `\textbf{+ CDL$^{-}$}' in which the weights $w_i$ of $\phi_i$ are simply computed based on the detection accuracy on the training data excluding $\mathcal{D}^s_i$ instead of using our model importance learning $\psi$.
Table~\ref{ab_1} illustrates the results under both settings using $M=10$, where the results for the hard setting are averaged over all the anomaly classes (see \texttt{Appendix C} for class-level results). We can observe that using only HADG, `\textbf{+ HADG}', largely improves DRA on most datasets, and it also substantially outperforms three types of DRA-based ensemble models, showing the effectiveness of our HADG component. Adding the CDL component, `\textbf{+ CDL}', consistently improves the variant `\textbf{+ HADG}' across all datasets. Simplifying `\textbf{+ CDL}' to `\textbf{+ CDL$^{-}$}' leads to performance drop on most datasets, with relatively large drops on challenging datasets like AITEX (both settings), Mastcam, and Hyper-Kvasir, indicating the significance of $\psi$ in CDL.

\begin{table}[t]
\resizebox{\linewidth}{!}{
\begin{tabular}{cc|cccc|cc|}
\hline
\multicolumn{1}{|c|}{\textbf{Dataset}}      & \textbf{DRA}  & \textbf{+ HADG}     & \textbf{+ RamHADG}   & \textbf{+ RamFULL}& \textbf{+ RamHADG + DevNet} & \textbf{+ CDL$^{-}$} & \textbf{+ CDL}            \\ 
\hline
\multicolumn{8}{|c|}{\textbf{General Setting} }                                                                                                    \\ \hline
\multicolumn{1}{|c|}{\textbf{AITEX}}     & \multicolumn{1}{c|}{0.892{\scriptsize±0.003}} & \multicolumn{1}{c}{\textbf{0.916{\scriptsize±0.004}}} & \multicolumn{1}{c}{0.908{\scriptsize±0.003}}  & \multicolumn{1}{c}{0.905{\scriptsize±0.011}} & \multicolumn{1}{c}{0.894{\scriptsize±0.003}}& \multicolumn{1}{|c}{0.915{\scriptsize±0.007}}& \multicolumn{1}{c|}{\color[HTML]{FF0000}{\textbf{0.925{\scriptsize±0.013}}}}\\
\multicolumn{1}{|c|}{\textbf{SDD}}       & \multicolumn{1}{c|}{\textbf{0.990{\scriptsize±0.000}}} & \multicolumn{1}{c}{\textbf{0.990{\scriptsize±0.000}}} & \multicolumn{1}{c}{\textbf{0.990{\scriptsize±0.000}}}  & \multicolumn{1}{c}{0.985{\scriptsize±0.002}} & \multicolumn{1}{c}{0.984{\scriptsize±0.000}}& \multicolumn{1}{|c}{\color[HTML]{FF0000}{\textbf{0.991{\scriptsize±0.001}}}}& \multicolumn{1}{c|}{\color[HTML]{FF0000}{\textbf{0.991{\scriptsize±0.000}}}} \\
\multicolumn{1}{|c|}{\textbf{ELPV}}      & \multicolumn{1}{c|}{0.843{\scriptsize±0.002}} & \multicolumn{1}{c}{0.846{\scriptsize±0.006}} & \multicolumn{1}{c}{0.844{\scriptsize±0.002}}  & \multicolumn{1}{c}{0.843{\scriptsize±0.000}} & \multicolumn{1}{c}{0.843{\scriptsize±0.003}}& \multicolumn{1}{|c}{\textbf{0.847{\scriptsize±0.002}}}& \multicolumn{1}{c|}{\textbf{0.850{\scriptsize±0.004}}} \\
\multicolumn{1}{|c|}{\textbf{optical}}   & \multicolumn{1}{c|}{0.966{\scriptsize±0.002}} & \multicolumn{1}{c}{\textbf{0.974{\scriptsize±0.002}}} & \multicolumn{1}{c}{0.970{\scriptsize±0.004}}  & \multicolumn{1}{c}{0.963{\scriptsize±0.006}} & \multicolumn{1}{c}{0.897{\scriptsize±0.005}}& \multicolumn{1}{|c}{\textbf{0.974{\scriptsize±0.002}}}& \multicolumn{1}{c|}{\color[HTML]{FF0000}{\textbf{0.976{\scriptsize±0.004}}}} \\
\multicolumn{1}{|c|}{\textbf{Mastcam}}   & \multicolumn{1}{c|}{0.849{\scriptsize±0.003}} & \multicolumn{1}{c}{\textbf{0.852{\scriptsize±0.001}}} & \multicolumn{1}{c}{0.815{\scriptsize±0.003}}  & \multicolumn{1}{c}{0.849{\scriptsize±0.001}} & \multicolumn{1}{c}{0.813{\scriptsize±0.001}}& \multicolumn{1}{|c}{0.841{\scriptsize±0.004}}& \multicolumn{1}{c|}{\color[HTML]{FF0000}{\textbf{0.855{\scriptsize±0.005}}}} \\
\multicolumn{1}{|c|}{\textbf{BrainMRI}}  & \multicolumn{1}{c|}{0.971{\scriptsize±0.001}} & \multicolumn{1}{c}{0.973{\scriptsize±0.002}} & \multicolumn{1}{c}{\textbf{0.974{\scriptsize±0.001}}}  & \multicolumn{1}{c}{0.973{\scriptsize±0.007}} & \multicolumn{1}{c}{0.961{\scriptsize±0.002}}& \multicolumn{1}{|c}{\color[HTML]{FF0000}{\textbf{0.977{\scriptsize±0.002}}}}& \multicolumn{1}{c|}{\color[HTML]{FF0000}{\textbf{0.977{\scriptsize±0.001}}}} \\
\multicolumn{1}{|c|}{\textbf{HeadCT}}    & \multicolumn{1}{c|}{0.978{\scriptsize±0.001}} & \multicolumn{1}{c}{0.992{\scriptsize±0.002}} & \multicolumn{1}{c}{0.986{\scriptsize±0.001}}  & \multicolumn{1}{c}{0.988{\scriptsize±0.003}} & \multicolumn{1}{c}{\color[HTML]{FF0000}{\textbf{0.995{\scriptsize±0.002}}}}& \multicolumn{1}{|c}{0.991{\scriptsize±0.001}}& \multicolumn{1}{c|}{\textbf{0.993{\scriptsize±0.002}}} \\
\multicolumn{1}{|c|}{\textbf{Hyper-Kvasir}}    & \multicolumn{1}{c|}{0.844{\scriptsize±0.001}} & \multicolumn{1}{c}{0.874{\scriptsize±0.005}} & \multicolumn{1}{c}{0.865{\scriptsize±0.002}} & \multicolumn{1}{c}{0.863{\scriptsize±0.001}} & \multicolumn{1}{c}{0.831{\scriptsize±0.003}}& \multicolumn{1}{|c}{\textbf{0.877{\scriptsize±0.005}}}& \multicolumn{1}{c|}{\color[HTML]{FF0000}{\textbf{0.880{\scriptsize±0.003}}}} \\
\multicolumn{1}{|c|}{\textbf{MVTecAD {\color[HTML]{808080} {\scriptsize(mean)}}}}   & \multicolumn{1}{c|}{0.966{\scriptsize±0.002}} & \multicolumn{1}{c}{\textbf{0.968{\scriptsize±0.003}}} & \multicolumn{1}{c}{0.964{\scriptsize±0.003}}  & \multicolumn{1}{c}{\textbf{0.968{\scriptsize±0.004}}} & \multicolumn{1}{c}{0.954{\scriptsize±0.005}}& \multicolumn{1}{|c}{0.966{\scriptsize±0.003}}& \multicolumn{1}{c|}{\color[HTML]{FF0000}{\textbf{0.970{\scriptsize±0.002}}}} \\
\hline

\multicolumn{8}{|c|}{\textbf{Hard Setting}}                                                                                                        \\ \hline
\multicolumn{1}{|c|}{\textbf{Carpet} {\color[HTML]{808080} {\scriptsize(mean)}}}   &   \multicolumn{1}{c|}{0.940{\scriptsize±0.006}}
& {0.943{\scriptsize±0.003}}          &   {0.943{\scriptsize±0.002}}       & \multicolumn{1}{c}{0.947{\scriptsize±0.001}} & \multicolumn{1}{c}{0.902{\scriptsize±0.002}}& \multicolumn{1}{|c}{\color[HTML]{FF0000}{\textbf{0.951{\scriptsize±0.007}}}}& \multicolumn{1}{c|}{\textbf{0.949{\scriptsize±0.002}}} \\ 

\multicolumn{1}{|c|}{\textbf{AITEX} {\color[HTML]{808080} {\scriptsize(mean)}}}  & \multicolumn{1}{c|}{0.733{\scriptsize±0.011}}   
& {\textbf{0.739{\scriptsize±0.007}}}         &    0.733{\scriptsize±0.005}    & \multicolumn{1}{c}{0.733{\scriptsize±0.004}} & \multicolumn{1}{c}{0.691{\scriptsize±0.005}}& \multicolumn{1}{|c}{0.720{\scriptsize±0.003}}& \multicolumn{1}{c|}{\color[HTML]{FF0000}{\textbf{0.747{\scriptsize±0.002}}}} \\ 

\multicolumn{1}{|c|}{\textbf{ELPV} {\color[HTML]{808080} {\scriptsize(mean)}}} & \multicolumn{1}{c|}{0.771{\scriptsize±0.005}}   
& {\textbf{0.784{\scriptsize±0.004}}}         &    0.774{\scriptsize±0.004}    & \multicolumn{1}{c}{0.779{\scriptsize±0.002}} & \multicolumn{1}{c}{0.732{\scriptsize±0.001}}& \multicolumn{1}{|c}{0.779{\scriptsize±0.009}}& \multicolumn{1}{c|}{\color[HTML]{FF0000}{\textbf{0.788{\scriptsize±0.003}}}} \\ 

\multicolumn{1}{|c|}{\textbf{Hyper-Kvasir} {\color[HTML]{808080} {\scriptsize(mean)}}} &  \multicolumn{1}{c|}{0.822{\scriptsize±0.013}}     
&  {\textbf{0.847{\scriptsize±0.008}}}         &   0.835{\scriptsize±0.004} & \multicolumn{1}{c}{0.829{\scriptsize±0.003}} & \multicolumn{1}{c}{0.833{\scriptsize±0.005}}& \multicolumn{1}{|c}{0.838{\scriptsize±0.011}}&   {\color[HTML]{FF0000}\textbf{0.854{\scriptsize±0.004}}}  \\ \hline

\end{tabular}}
\caption{Ablation study results of \coolname and its variants under both general and hard settings. Best results and the second-best results are respectively highlighted in {\color[HTML]{FF0000} \textbf{red}} and \textbf{bold}.
}
\label{ab_1}
\vspace{-0.5cm}
\end{table}

\subsubsection{Hyperparameter Analysis}
\begin{figure}[t]
\centering
\includegraphics[width=0.23\textwidth]{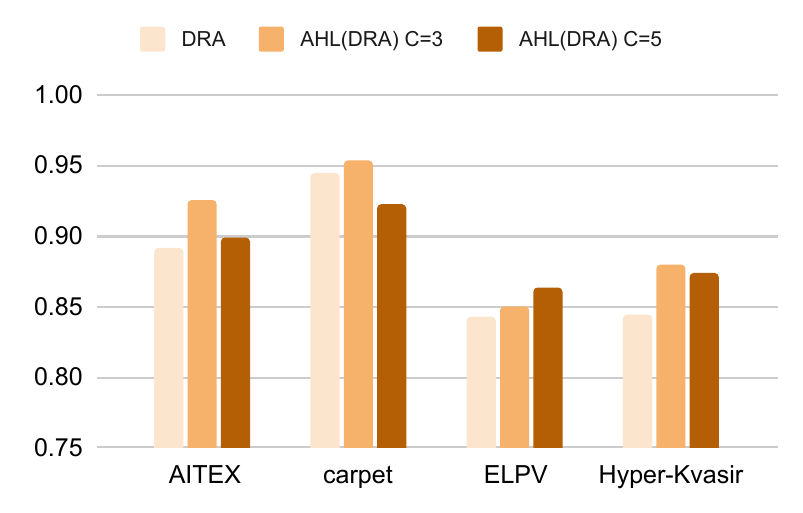}
\includegraphics[width=0.23\textwidth]{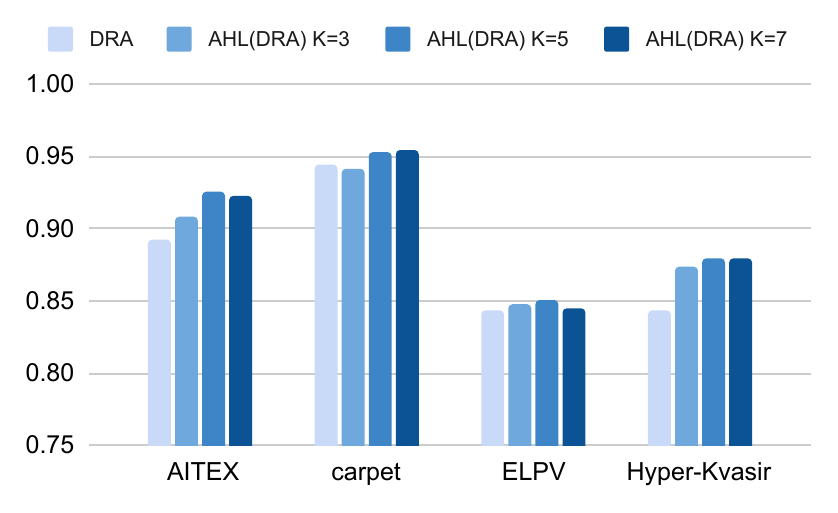}
\caption{Hyperparameter analysis of \coolname (DRA) based on the general setting using ten anomaly examples. \textbf{Left:} AUC results w.r.t. different number of clusters ($C$). \textbf{Right:} AUC results w.r.t. the length of sequences ($K$) as input to the sequential model $\psi$. }
\label{hyper}
\vspace{-0.2cm}
\end{figure}

We conduct an analysis of two key hyperparameters in \coolname, the number of normal clusters ($C$) in HADG (Sec. \ref{subsec:hadg}) and the length of score sequences ($K$) in CDL (Sec. \ref{subsec:cdl}), with the results shown in Fig.~\ref{hyper} (left) and Fig.~\ref{hyper} (right), respectively.
We observe that increasing $C$ does not always result in improved performance. This is mainly because that excessively dividing normal samples into too many small clusters can introduce biased individual anomaly distributions into the learning process due to the presence of too few samples per cluster. 
As for $K$, the results show that setting $K=3$ is generally sufficient for accurate generalization error estimation. However, having $K=5$ can further improve the performance on a few datasets where more accurate generalization error estimation requires longer input sequences.
$C=3$ and $K=5$ are generally suggested, which are the default settings for \coolname models throughout our large-scale experiments.

\section{Conclusion}
\label{sec:conclusion}

In this work, we explore the OSAD problem and introduce a novel, generic framework, namely anomaly heterogeneity learning (\coolname). 
It learns generalized, heterogeneous abnormality detection capability by training on diverse generated anomaly distributions in simulated OSAD scenarios. \coolname models such anomaly heterogeneity using a collaborative differentiable learning on a set of heterogeneous based models built on the generated anomaly distribution.
Experiments on nine real-world anomaly detection datasets demonstrate that the \coolname approach can substantially enhance different state-of-the-art OSAD models
in detecting unseen anomalies in the same-domain and cross-domain cases, with the AUC improvement up to 9\%.

{
    \small
    \bibliographystyle{ieeenat_fullname}
    \bibliography{main}

\begin{thebibliography}{68}
\providecommand{\natexlab}[1]{#1}
\providecommand{\url}[1]{\texttt{#1}}
\expandafter\ifx\csname urlstyle\endcsname\relax
  \providecommand{\doi}[1]{doi: #1}\else
  \providecommand{\doi}{doi: \begingroup \urlstyle{rm}\Url}\fi

\bibitem[Acsintoae et~al.(2022)Acsintoae, Florescu, Georgescu, Mare, Sumedrea, Ionescu, Khan, and Shah]{acsintoae2022ubnormal}
Andra Acsintoae, Andrei Florescu, Mariana-Iuliana Georgescu, Tudor Mare, Paul Sumedrea, Radu~Tudor Ionescu, Fahad~Shahbaz Khan, and Mubarak Shah.
\newblock Ubnormal: New benchmark for supervised open-set video anomaly detection.
\newblock In \emph{Proceedings of the IEEE/CVF Conference on Computer Vision and Pattern Recognition}, pages 20143--20153, 2022.

\bibitem[Akcay et~al.(2019)Akcay, Atapour-Abarghouei, and Breckon]{akcay2019ganomaly}
Samet Akcay, Amir Atapour-Abarghouei, and Toby~P Breckon.
\newblock Ganomaly: Semi-supervised anomaly detection via adversarial training.
\newblock In \emph{Computer Vision--ACCV 2018: 14th Asian Conference on Computer Vision, Perth, Australia, December 2--6, 2018, Revised Selected Papers, Part III 14}, pages 622--637. Springer, 2019.

\bibitem[Astrid et~al.(2023)Astrid, Zaheer, and Lee]{astrid2023pseudobound}
Marcella Astrid, Muhammad~Zaigham Zaheer, and Seung-Ik Lee.
\newblock Pseudobound: Limiting the anomaly reconstruction capability of one-class classifiers using pseudo anomalies.
\newblock \emph{Neurocomputing}, 534:\penalty0 147--160, 2023.

\bibitem[Bergman and Hoshen(2020)]{bergman2020classification}
Liron Bergman and Yedid Hoshen.
\newblock Classification-based anomaly detection for general data.
\newblock \emph{arXiv preprint arXiv:2005.02359}, 2020.

\bibitem[Bergmann et~al.(2019)Bergmann, Fauser, Sattlegger, and Steger]{bergmann2019mvtec}
Paul Bergmann, Michael Fauser, David Sattlegger, and Carsten Steger.
\newblock Mvtec ad--a comprehensive real-world dataset for unsupervised anomaly detection.
\newblock In \emph{Proceedings of the IEEE/CVF conference on computer vision and pattern recognition}, pages 9592--9600, 2019.

\bibitem[Bergmann et~al.(2020)Bergmann, Fauser, Sattlegger, and Steger]{bergmann2020uninformed}
Paul Bergmann, Michael Fauser, David Sattlegger, and Carsten Steger.
\newblock Uninformed students: Student-teacher anomaly detection with discriminative latent embeddings.
\newblock In \emph{Proceedings of the IEEE/CVF conference on computer vision and pattern recognition}, pages 4183--4192, 2020.

\bibitem[Borgli et~al.(2020)Borgli, Thambawita, Smedsrud, Hicks, Jha, Eskeland, Randel, Pogorelov, Lux, Nguyen, et~al.]{borgli2020hyperkvasir}
Hanna Borgli, Vajira Thambawita, Pia~H Smedsrud, Steven Hicks, Debesh Jha, Sigrun~L Eskeland, Kristin~Ranheim Randel, Konstantin Pogorelov, Mathias Lux, Duc Tien~Dang Nguyen, et~al.
\newblock Hyperkvasir, a comprehensive multi-class image and video dataset for gastrointestinal endoscopy.
\newblock \emph{Scientific data}, 7\penalty0 (1):\penalty0 283, 2020.

\bibitem[Bozic et~al.(2021)Bozic, Tabernik, and Skocaj]{BozicTS21}
Jakob Bozic, Domen Tabernik, and Danijel Skocaj.
\newblock Mixed supervision for surface-defect detection: From weakly to fully supervised learning.
\newblock \emph{Comput. Ind.}, 129:\penalty0 103459, 2021.

\bibitem[Cao et~al.(2023)Cao, Zhu, and Pang]{cao2023anomaly}
Tri Cao, Jiawen Zhu, and Guansong Pang.
\newblock Anomaly detection under distribution shift.
\newblock In \emph{Proceedings of the IEEE/CVF International Conference on Computer Vision (ICCV)}, pages 6511--6523, 2023.

\bibitem[Chen et~al.(2022)Chen, Tian, Pang, and Carneiro]{chen2022deep}
Yuanhong Chen, Yu Tian, Guansong Pang, and Gustavo Carneiro.
\newblock Deep one-class classification via interpolated gaussian descriptor.
\newblock In \emph{Proceedings of the AAAI Conference on Artificial Intelligence}, pages 383--392, 2022.

\bibitem[Chen et~al.(2023)Chen, Liu, Zhang, Fok, Qi, and Wu]{chen2023mgfn}
Yingxian Chen, Zhengzhe Liu, Baoheng Zhang, Wilton Fok, Xiaojuan Qi, and Yik-Chung Wu.
\newblock Mgfn: Magnitude-contrastive glance-and-focus network for weakly-supervised video anomaly detection.
\newblock In \emph{Proceedings of the AAAI Conference on Artificial Intelligence}, pages 387--395, 2023.

\bibitem[Chu and Kitani(2020)]{ChuK20}
Wen{-}Hsuan Chu and Kris~M. Kitani.
\newblock Neural batch sampling with reinforcement learning for semi-supervised anomaly detection.
\newblock In \emph{Computer Vision - {ECCV} 2020 - 16th European Conference, Glasgow, UK, August 23-28, 2020, Proceedings, Part {XXVI}}, pages 751--766. Springer, 2020.

\bibitem[Deitsch et~al.(2019)Deitsch, Christlein, Berger, Buerhop-Lutz, Maier, Gallwitz, and Riess]{deitsch2019automatic}
Sergiu Deitsch, Vincent Christlein, Stephan Berger, Claudia Buerhop-Lutz, Andreas Maier, Florian Gallwitz, and Christian Riess.
\newblock Automatic classification of defective photovoltaic module cells in electroluminescence images.
\newblock \emph{Solar Energy}, 185:\penalty0 455--468, 2019.

\bibitem[Deng and Li(2022)]{deng2022anomaly}
Hanqiu Deng and Xingyu Li.
\newblock Anomaly detection via reverse distillation from one-class embedding.
\newblock In \emph{Proceedings of the IEEE/CVF Conference on Computer Vision and Pattern Recognition}, pages 9737--9746, 2022.

\bibitem[Ding et~al.(2022)Ding, Pang, and Shen]{ding2022catching}
Choubo Ding, Guansong Pang, and Chunhua Shen.
\newblock Catching both gray and black swans: Open-set supervised anomaly detection.
\newblock In \emph{Proceedings of the IEEE/CVF Conference on Computer Vision and Pattern Recognition}, pages 7388--7398, 2022.

\bibitem[Georgescu et~al.(2021)Georgescu, Barbalau, Ionescu, Khan, Popescu, and Shah]{georgescu2021anomaly}
Mariana-Iuliana Georgescu, Antonio Barbalau, Radu~Tudor Ionescu, Fahad~Shahbaz Khan, Marius Popescu, and Mubarak Shah.
\newblock Anomaly detection in video via self-supervised and multi-task learning.
\newblock In \emph{Proceedings of the IEEE/CVF conference on computer vision and pattern recognition}, pages 12742--12752, 2021.

\bibitem[Goodfellow et~al.(2020)Goodfellow, Pouget-Abadie, Mirza, Xu, Warde-Farley, Ozair, Courville, and Bengio]{goodfellow2020generative}
Ian Goodfellow, Jean Pouget-Abadie, Mehdi Mirza, Bing Xu, David Warde-Farley, Sherjil Ozair, Aaron Courville, and Yoshua Bengio.
\newblock Generative adversarial networks.
\newblock \emph{Communications of the ACM}, 63\penalty0 (11):\penalty0 139--144, 2020.

\bibitem[Hou et~al.(2021)Hou, Zhang, Zhong, Xie, Pu, and Zhou]{hou2021divide}
Jinlei Hou, Yingying Zhang, Qiaoyong Zhong, Di Xie, Shiliang Pu, and Hong Zhou.
\newblock Divide-and-assemble: Learning block-wise memory for unsupervised anomaly detection.
\newblock In \emph{Proceedings of the IEEE/CVF International Conference on Computer Vision}, pages 8791--8800, 2021.

\bibitem[Huang et~al.(2022)Huang, Xu, Wang, Wang, and Zhang]{huang2022self}
Chaoqin Huang, Qinwei Xu, Yanfeng Wang, Yu Wang, and Ya Zhang.
\newblock Self-supervised masking for unsupervised anomaly detection and localization.
\newblock \emph{IEEE Transactions on Multimedia}, 2022.

\bibitem[Kerner et~al.(2020)Kerner, Wagstaff, Bue, Wellington, Jacob, Horton, Bell, Kwan, and Ben~Amor]{kerner2020comparison}
Hannah~R Kerner, Kiri~L Wagstaff, Brian~D Bue, Danika~F Wellington, Samantha Jacob, Paul Horton, James~F Bell, Chiman Kwan, and Heni Ben~Amor.
\newblock Comparison of novelty detection methods for multispectral images in rover-based planetary exploration missions.
\newblock \emph{Data Mining and Knowledge Discovery}, 34:\penalty0 1642--1675, 2020.

\bibitem[Kingma and Welling(2013)]{kingma2013auto}
Diederik~P Kingma and Max Welling.
\newblock Auto-encoding variational bayes.
\newblock \emph{arXiv preprint arXiv:1312.6114}, 2013.

\bibitem[Li et~al.(2021)Li, Sohn, Yoon, and Pfister]{li2021cutpaste}
Chun-Liang Li, Kihyuk Sohn, Jinsung Yoon, and Tomas Pfister.
\newblock Cutpaste: Self-supervised learning for anomaly detection and localization.
\newblock In \emph{Proceedings of the IEEE/CVF Conference on Computer Vision and Pattern Recognition}, pages 9664--9674, 2021.

\bibitem[Lin et~al.(2017)Lin, Goyal, Girshick, He, and Doll{\'a}r]{lin2017focal}
Tsung-Yi Lin, Priya Goyal, Ross Girshick, Kaiming He, and Piotr Doll{\'a}r.
\newblock Focal loss for dense object detection.
\newblock In \emph{Proceedings of the IEEE international conference on computer vision}, pages 2980--2988, 2017.

\bibitem[Liu et~al.(2019)Liu, Luo, Li, Zhao, Gao, et~al.]{liu2019margin}
Wen Liu, Weixin Luo, Zhengxin Li, Peilin Zhao, Shenghua Gao, et~al.
\newblock Margin learning embedded prediction for video anomaly detection with a few anomalies.
\newblock In \emph{IJCAI}, pages 3023--3030, 2019.

\bibitem[Liu et~al.(2023{\natexlab{a}})Liu, Chang, Ma, Shan, and Chen]{liu2023diversity}
Wenrui Liu, Hong Chang, Bingpeng Ma, Shiguang Shan, and Xilin Chen.
\newblock Diversity-measurable anomaly detection.
\newblock In \emph{Proceedings of the IEEE/CVF Conference on Computer Vision and Pattern Recognition}, pages 12147--12156, 2023{\natexlab{a}}.

\bibitem[Liu et~al.(2023{\natexlab{b}})Liu, Zhou, Xu, and Wang]{liu2023simplenet}
Zhikang Liu, Yiming Zhou, Yuansheng Xu, and Zilei Wang.
\newblock Simplenet: A simple network for image anomaly detection and localization, 2023{\natexlab{b}}.

\bibitem[Liznerski et~al.(2020)Liznerski, Ruff, Vandermeulen, Franks, Kloft, and M{\"{u}}ller]{abs-2007-01760}
Philipp Liznerski, Lukas Ruff, Robert~A. Vandermeulen, Billy~Joe Franks, Marius Kloft, and Klaus{-}Robert M{\"{u}}ller.
\newblock Explainable deep one-class classification.
\newblock \emph{CoRR}, abs/2007.01760, 2020.

\bibitem[Lu et~al.(2020)Lu, Yu, Reddy, and Wang]{LuYR020}
Yiwei Lu, Frank Yu, Mahesh Kumar~Krishna Reddy, and Yang Wang.
\newblock Few-shot scene-adaptive anomaly detection.
\newblock In \emph{Computer Vision - {ECCV} 2020 - 16th European Conference, Glasgow, UK, August 23-28, 2020, Proceedings, Part {V}}, pages 125--141. Springer, 2020.

\bibitem[Lv et~al.(2023)Lv, Yue, Sun, Luo, Cui, and Zhang]{lv2023unbiased}
Hui Lv, Zhongqi Yue, Qianru Sun, Bin Luo, Zhen Cui, and Hanwang Zhang.
\newblock Unbiased multiple instance learning for weakly supervised video anomaly detection.
\newblock In \emph{Proceedings of the IEEE/CVF Conference on Computer Vision and Pattern Recognition}, pages 8022--8031, 2023.

\bibitem[Markovitz et~al.(2020)Markovitz, Sharir, Friedman, Zelnik-Manor, and Avidan]{markovitz2020graph}
Amir Markovitz, Gilad Sharir, Itamar Friedman, Lihi Zelnik-Manor, and Shai Avidan.
\newblock Graph embedded pose clustering for anomaly detection.
\newblock In \emph{Proceedings of the IEEE/CVF Conference on Computer Vision and Pattern Recognition}, pages 10539--10547, 2020.

\bibitem[Pang et~al.(2018)Pang, Cao, Chen, and Liu]{pang2018learning}
Guansong Pang, Longbing Cao, Ling Chen, and Huan Liu.
\newblock Learning representations of ultrahigh-dimensional data for random distance-based outlier detection.
\newblock In \emph{Proceedings of the 24th ACM SIGKDD international conference on knowledge discovery \& data mining}, pages 2041--2050, 2018.

\bibitem[Pang et~al.(2021{\natexlab{a}})Pang, Ding, Shen, and Hengel]{pang2021explainable}
Guansong Pang, Choubo Ding, Chunhua Shen, and Anton van~den Hengel.
\newblock Explainable deep few-shot anomaly detection with deviation networks.
\newblock \emph{arXiv preprint arXiv:2108.00462}, 2021{\natexlab{a}}.

\bibitem[Pang et~al.(2021{\natexlab{b}})Pang, Shen, Cao, and Hengel]{pang2021deep}
Guansong Pang, Chunhua Shen, Longbing Cao, and Anton Van~Den Hengel.
\newblock Deep learning for anomaly detection: A review.
\newblock \emph{ACM computing surveys (CSUR)}, 54\penalty0 (2):\penalty0 1--38, 2021{\natexlab{b}}.

\bibitem[Pang et~al.(2021{\natexlab{c}})Pang, van~den Hengel, Shen, and Cao]{pang2021toward}
Guansong Pang, Anton van~den Hengel, Chunhua Shen, and Longbing Cao.
\newblock Toward deep supervised anomaly detection: Reinforcement learning from partially labeled anomaly data.
\newblock In \emph{Proceedings of the 27th ACM SIGKDD conference on knowledge discovery \& data mining}, pages 1298--1308, 2021{\natexlab{c}}.

\bibitem[Pang et~al.(2023)Pang, Shen, Jin, and van~den Hengel]{pang2023deep}
Guansong Pang, Chunhua Shen, Huidong Jin, and Anton van~den Hengel.
\newblock Deep weakly-supervised anomaly detection.
\newblock In \emph{Proceedings of the 29th ACM SIGKDD Conference on Knowledge Discovery and Data Mining}, pages 1795--1807, 2023.

\bibitem[Park et~al.(2020)Park, Noh, and Ham]{park2020learning}
Hyunjong Park, Jongyoun Noh, and Bumsub Ham.
\newblock Learning memory-guided normality for anomaly detection.
\newblock In \emph{Proceedings of the IEEE/CVF conference on computer vision and pattern recognition}, pages 14372--14381, 2020.

\bibitem[Ristea et~al.(2022)Ristea, Madan, Ionescu, Nasrollahi, Khan, Moeslund, and Shah]{ristea2022self}
Nicolae-C{\u{a}}t{\u{a}}lin Ristea, Neelu Madan, Radu~Tudor Ionescu, Kamal Nasrollahi, Fahad~Shahbaz Khan, Thomas~B Moeslund, and Mubarak Shah.
\newblock Self-supervised predictive convolutional attentive block for anomaly detection.
\newblock In \emph{Proceedings of the IEEE/CVF Conference on Computer Vision and Pattern Recognition}, pages 13576--13586, 2022.

\bibitem[Roth et~al.(2022)Roth, Pemula, Zepeda, Sch{\"o}lkopf, Brox, and Gehler]{roth2022towards}
Karsten Roth, Latha Pemula, Joaquin Zepeda, Bernhard Sch{\"o}lkopf, Thomas Brox, and Peter Gehler.
\newblock Towards total recall in industrial anomaly detection.
\newblock In \emph{Proceedings of the IEEE/CVF Conference on Computer Vision and Pattern Recognition}, pages 14318--14328, 2022.

\bibitem[Ruff et~al.(2020)Ruff, Vandermeulen, G{\"o}rnitz, Binder, M{\"u}ller, M{\"u}ller, and Kloft]{ruff2020deep}
Lukas Ruff, Robert~A Vandermeulen, Nico G{\"o}rnitz, Alexander Binder, Emmanuel M{\"u}ller, Klaus-Robert M{\"u}ller, and Marius Kloft.
\newblock Deep semi-supervised anomaly detection.
\newblock In \emph{ICLR}, 2020.

\bibitem[Salehi et~al.(2021)Salehi, Sadjadi, Baselizadeh, Rohban, and Rabiee]{salehi2021multiresolution}
Mohammadreza Salehi, Niousha Sadjadi, Soroosh Baselizadeh, Mohammad~H Rohban, and Hamid~R Rabiee.
\newblock Multiresolution knowledge distillation for anomaly detection.
\newblock In \emph{Proceedings of the IEEE/CVF conference on computer vision and pattern recognition}, pages 14902--14912, 2021.

\bibitem[Schlegl et~al.(2019)Schlegl, Seeb{\"o}ck, Waldstein, Langs, and Schmidt-Erfurth]{schlegl2019f}
Thomas Schlegl, Philipp Seeb{\"o}ck, Sebastian~M Waldstein, Georg Langs, and Ursula Schmidt-Erfurth.
\newblock f-anogan: Fast unsupervised anomaly detection with generative adversarial networks.
\newblock \emph{Medical image analysis}, 54:\penalty0 30--44, 2019.

\bibitem[Silvestre-Blanes et~al.(2019)Silvestre-Blanes, Albero-Albero, Miralles, P{\'e}rez-Llorens, and Moreno]{silvestre2019public}
Javier Silvestre-Blanes, Teresa Albero-Albero, Ignacio Miralles, Rub{\'e}n P{\'e}rez-Llorens, and Jorge Moreno.
\newblock A public fabric database for defect detection methods and results.
\newblock \emph{Autex Research Journal}, 19\penalty0 (4):\penalty0 363--374, 2019.

\bibitem[Sultani et~al.(2018)Sultani, Chen, and Shah]{sultani2018real}
Waqas Sultani, Chen Chen, and Mubarak Shah.
\newblock Real-world anomaly detection in surveillance videos.
\newblock In \emph{Proceedings of the IEEE conference on computer vision and pattern recognition}, pages 6479--6488, 2018.

\bibitem[Tabernik et~al.(2020)Tabernik, {\v{S}}ela, Skvar{\v{c}}, and Sko{\v{c}}aj]{tabernik2020segmentation}
Domen Tabernik, Samo {\v{S}}ela, Jure Skvar{\v{c}}, and Danijel Sko{\v{c}}aj.
\newblock Segmentation-based deep-learning approach for surface-defect detection.
\newblock \emph{Journal of Intelligent Manufacturing}, 31\penalty0 (3):\penalty0 759--776, 2020.

\bibitem[Tack et~al.(2020)Tack, Mo, Jeong, and Shin]{tack2020csi}
Jihoon Tack, Sangwoo Mo, Jongheon Jeong, and Jinwoo Shin.
\newblock Csi: Novelty detection via contrastive learning on distributionally shifted instances.
\newblock \emph{Advances in neural information processing systems}, 33:\penalty0 11839--11852, 2020.

\bibitem[Tax and Duin(2004)]{tax2004support}
David~MJ Tax and Robert~PW Duin.
\newblock Support vector data description.
\newblock \emph{Machine learning}, 54:\penalty0 45--66, 2004.

\bibitem[Tian et~al.(2021)Tian, Pang, Chen, Singh, Verjans, and Carneiro]{tian2021weakly}
Yu Tian, Guansong Pang, Yuanhong Chen, Rajvinder Singh, Johan~W Verjans, and Gustavo Carneiro.
\newblock Weakly-supervised video anomaly detection with robust temporal feature magnitude learning.
\newblock In \emph{Proceedings of the IEEE/CVF international conference on computer vision}, pages 4975--4986, 2021.

\bibitem[Tien et~al.(2023)Tien, Nguyen, Tran, Huy, Duong, Nguyen, and Truong]{tien2023revisiting}
Tran~Dinh Tien, Anh~Tuan Nguyen, Nguyen~Hoang Tran, Ta~Duc Huy, Soan Duong, Chanh D~Tr Nguyen, and Steven~QH Truong.
\newblock Revisiting reverse distillation for anomaly detection.
\newblock In \emph{Proceedings of the IEEE/CVF Conference on Computer Vision and Pattern Recognition}, pages 24511--24520, 2023.

\bibitem[Wang et~al.(2021)Wang, Han, Ding, and Huang]{wang2021student}
Guodong Wang, Shumin Han, Errui Ding, and Di Huang.
\newblock Student-teacher feature pyramid matching for anomaly detection.
\newblock \emph{arXiv preprint arXiv:2103.04257}, 2021.

\bibitem[Wieler and Hahn(2007)]{wieler2007weakly}
Matthias Wieler and Tobias Hahn.
\newblock Weakly supervised learning for industrial optical inspection.
\newblock In \emph{DAGM symposium in}, 2007.

\bibitem[Wu et~al.(2021)Wu, Chen, Fuh, and Liu]{WuCFL21}
Jhih{-}Ciang Wu, Ding{-}Jie Chen, Chiou{-}Shann Fuh, and Tyng{-}Luh Liu.
\newblock Learning unsupervised metaformer for anomaly detection.
\newblock In \emph{2021 {IEEE/CVF} International Conference on Computer Vision, {ICCV} 2021, Montreal, QC, Canada, October 10-17, 2021}, pages 4349--4358. {IEEE}, 2021.

\bibitem[Wu et~al.(2020)Wu, Liu, Shi, Sun, Shao, Wu, and Yang]{wu2020not}
Peng Wu, Jing Liu, Yujia Shi, Yujia Sun, Fangtao Shao, Zhaoyang Wu, and Zhiwei Yang.
\newblock Not only look, but also listen: Learning multimodal violence detection under weak supervision.
\newblock In \emph{Computer Vision--ECCV 2020: 16th European Conference, Glasgow, UK, August 23--28, 2020, Proceedings, Part XXX 16}, pages 322--339. Springer, 2020.

\bibitem[Wu et~al.(2023)Wu, Zhou, Pang, Zhou, Yan, Wang, and Zhang]{wu2023vadclip}
Peng Wu, Xuerong Zhou, Guansong Pang, Lingru Zhou, Qingsen Yan, Peng Wang, and Yanning Zhang.
\newblock Vadclip: Adapting vision-language models for weakly supervised video anomaly detection.
\newblock \emph{arXiv preprint arXiv:2308.11681}, 2023.

\bibitem[Xiang et~al.(2023)Xiang, Zhang, Lu, Yuille, Zhang, Cai, and Zhou]{xiang2023squid}
Tiange Xiang, Yixiao Zhang, Yongyi Lu, Alan~L Yuille, Chaoyi Zhang, Weidong Cai, and Zongwei Zhou.
\newblock Squid: Deep feature in-painting for unsupervised anomaly detection.
\newblock In \emph{Proceedings of the IEEE/CVF Conference on Computer Vision and Pattern Recognition}, pages 23890--23901, 2023.

\bibitem[Yan et~al.(2021)Yan, Zhang, Xu, Hu, and Heng]{yan2021learning}
Xudong Yan, Huaidong Zhang, Xuemiao Xu, Xiaowei Hu, and Pheng-Ann Heng.
\newblock Learning semantic context from normal samples for unsupervised anomaly detection.
\newblock In \emph{Proceedings of the AAAI Conference on Artificial Intelligence}, pages 3110--3118, 2021.

\bibitem[Yao et~al.(2022)Yao, Zhang, and Finn]{YaoZF22}
Huaxiu Yao, Linjun Zhang, and Chelsea Finn.
\newblock Meta-learning with fewer tasks through task interpolation.
\newblock In \emph{The Tenth International Conference on Learning Representations, {ICLR} 2022, Virtual Event, April 25-29, 2022}. OpenReview.net, 2022.

\bibitem[Yao et~al.(2023{\natexlab{a}})Yao, Li, Qian, Luo, and Zhang]{FOD}
Xincheng Yao, Ruoqi Li, Zefeng Qian, Yan Luo, and Chongyang Zhang.
\newblock Focus the discrepancy: Intra- and inter-correlation learning for image anomaly detection.
\newblock 2023{\natexlab{a}}.

\bibitem[Yao et~al.(2023{\natexlab{b}})Yao, Li, Zhang, Sun, and Zhang]{yao2023explicit}
Xincheng Yao, Ruoqi Li, Jing Zhang, Jun Sun, and Chongyang Zhang.
\newblock Explicit boundary guided semi-push-pull contrastive learning for supervised anomaly detection.
\newblock In \emph{Proceedings of the IEEE/CVF Conference on Computer Vision and Pattern Recognition}, pages 24490--24499, 2023{\natexlab{b}}.

\bibitem[Yi and Yoon(2020)]{yi2020patch}
Jihun Yi and Sungroh Yoon.
\newblock Patch svdd: Patch-level svdd for anomaly detection and segmentation.
\newblock In \emph{Proceedings of the Asian Conference on Computer Vision}, 2020.

\bibitem[Yun et~al.(2019)Yun, Han, Oh, Chun, Choe, and Yoo]{yun2019cutmix}
Sangdoo Yun, Dongyoon Han, Seong~Joon Oh, Sanghyuk Chun, Junsuk Choe, and Youngjoon Yoo.
\newblock Cutmix: Regularization strategy to train strong classifiers with localizable features.
\newblock In \emph{Proceedings of the IEEE/CVF international conference on computer vision}, pages 6023--6032, 2019.

\bibitem[Zaheer et~al.(2020)Zaheer, Lee, Astrid, and Lee]{zaheer2020old}
Muhammad~Zaigham Zaheer, Jin-ha Lee, Marcella Astrid, and Seung-Ik Lee.
\newblock Old is gold: Redefining the adversarially learned one-class classifier training paradigm.
\newblock In \emph{Proceedings of the IEEE/CVF Conference on Computer Vision and Pattern Recognition}, pages 14183--14193, 2020.

\bibitem[Zaheer et~al.(2022)Zaheer, Mahmood, Khan, Segu, Yu, and Lee]{zaheer2022generative}
M~Zaigham Zaheer, Arif Mahmood, M~Haris Khan, Mattia Segu, Fisher Yu, and Seung-Ik Lee.
\newblock Generative cooperative learning for unsupervised video anomaly detection.
\newblock In \emph{Proceedings of the IEEE/CVF Conference on Computer Vision and Pattern Recognition}, pages 14744--14754, 2022.

\bibitem[Zavrtanik et~al.(2021{\natexlab{a}})Zavrtanik, Kristan, and Sko{\v{c}}aj]{zavrtanik2021draem}
Vitjan Zavrtanik, Matej Kristan, and Danijel Sko{\v{c}}aj.
\newblock Draem-a discriminatively trained reconstruction embedding for surface anomaly detection.
\newblock In \emph{Proceedings of the IEEE/CVF International Conference on Computer Vision}, pages 8330--8339, 2021{\natexlab{a}}.

\bibitem[Zavrtanik et~al.(2021{\natexlab{b}})Zavrtanik, Kristan, and Sko{\v{c}}aj]{zavrtanik2021reconstruction}
Vitjan Zavrtanik, Matej Kristan, and Danijel Sko{\v{c}}aj.
\newblock Reconstruction by inpainting for visual anomaly detection.
\newblock \emph{Pattern Recognition}, 112:\penalty0 107706, 2021{\natexlab{b}}.

\bibitem[Zhang et~al.(2023{\natexlab{a}})Zhang, Wu, Wang, Chen, and Jiang]{zhang2023prototypical}
Hui Zhang, Zuxuan Wu, Zheng Wang, Zhineng Chen, and Yu-Gang Jiang.
\newblock Prototypical residual networks for anomaly detection and localization.
\newblock In \emph{Proceedings of the IEEE/CVF Conference on Computer Vision and Pattern Recognition}, pages 16281--16291, 2023{\natexlab{a}}.

\bibitem[Zhang et~al.(2023{\natexlab{b}})Zhang, Li, Li, Huang, Shan, and Chen]{zhang2023destseg}
Xuan Zhang, Shiyu Li, Xi Li, Ping Huang, Jiulong Shan, and Ting Chen.
\newblock Destseg: Segmentation guided denoising student-teacher for anomaly detection, 2023{\natexlab{b}}.

\bibitem[Zhou et~al.(2016)Zhou, Shi, Tian, Qi, Li, Hao, and Xu]{zhou2016attention}
Peng Zhou, Wei Shi, Jun Tian, Zhenyu Qi, Bingchen Li, Hongwei Hao, and Bo Xu.
\newblock Attention-based bidirectional long short-term memory networks for relation classification.
\newblock In \emph{Proceedings of the 54th annual meeting of the association for computational linguistics (volume 2: Short papers)}, pages 207--212, 2016.

\bibitem[Zhu et~al.(2022)Zhu, Bao, and Yu]{zhu2022towards}
Yuansheng Zhu, Wentao Bao, and Qi Yu.
\newblock Towards open set video anomaly detection.
\newblock In \emph{Computer Vision--ECCV 2022: 17th European Conference, Tel Aviv, Israel, October 23--27, 2022, Proceedings, Part XXXIV}, pages 395--412. Springer, 2022.

\end{thebibliography}
}

\newpage
\appendix

\section{Dataset Details}
\subsection{Key Data Statistics}
We conduct extensive experiments on nine real-world Anomaly Detection (AD) datasets. Table~\ref{datset detail} provides key data statistics for all the datasets used in this study. We follow exactly the same settings as prior Open-set Supervised Anomaly Detection (OSAD) studies \cite{pang2021explainable,ding2022catching}. Particularly, we follow the original settings of MVTec AD and split the normal samples into training and test sets; for the other eight datasets, the normal samples are randomly split into training and test sets using a 3:1 proportion.

\begin{table}[ht]
\centering
\resizebox{0.75\linewidth}{!}{
\begin{tabular}{|ccc|c|cc|}
\hline
\multicolumn{3}{|c|}{\textbf{Dataset}}                                                          & \textbf{Original Training} & \multicolumn{2}{c|}{\textbf{Original Test}} \\ \hline
\multicolumn{1}{|c|}{\textbf{}}             & \multicolumn{1}{c|}{\textbf{$|C|$}} & \textbf{Type} & \textbf{Normal}            & \textbf{Normal}      & \textbf{Anomaly}     \\ \hline
\multicolumn{1}{|c|}{\textbf{Carpet}}       & \multicolumn{1}{c|}{5}            & Texture       & 280                        & 28                   & 89                   \\
\multicolumn{1}{|c|}{\textbf{Grid}}         & \multicolumn{1}{c|}{5}            & Texture       & 264                        & 21                   & 57                   \\
\multicolumn{1}{|c|}{\textbf{Leather}}      & \multicolumn{1}{c|}{5}            & Texture       & 245                        & 32                   & 92                   \\
\multicolumn{1}{|c|}{\textbf{Tile}}         & \multicolumn{1}{c|}{5}            & Texture       & 230                        & 33                   & 83                   \\
\multicolumn{1}{|c|}{\textbf{Wood}}         & \multicolumn{1}{c|}{5}            & Texture       & 247                        & 19                   & 60                   \\
\multicolumn{1}{|c|}{\textbf{Bottle}}       & \multicolumn{1}{c|}{3}            & Object        & 209                        & 20                   & 63                   \\
\multicolumn{1}{|c|}{\textbf{Capsule}}      & \multicolumn{1}{c|}{5}            & Object        & 219                        & 23                   & 109                  \\
\multicolumn{1}{|c|}{\textbf{Pill}}         & \multicolumn{1}{c|}{7}            & Object        & 267                        & 26                   & 141                  \\
\multicolumn{1}{|c|}{\textbf{Transistor}}   & \multicolumn{1}{c|}{4}            & Object        & 213                        & 60                   & 40                   \\
\multicolumn{1}{|c|}{\textbf{Zipper}}       & \multicolumn{1}{c|}{7}            & Object        & 240                        & 32                   & 119                  \\
\multicolumn{1}{|c|}{\textbf{Cable}}        & \multicolumn{1}{c|}{8}            & Object        & 224                        & 58                   & 92                   \\
\multicolumn{1}{|c|}{\textbf{Hazelnut}}     & \multicolumn{1}{c|}{4}            & Object        & 391                        & 40                   & 70                   \\
\multicolumn{1}{|c|}{\textbf{Metal\_nut}}   & \multicolumn{1}{c|}{4}            & Object        & 220                        & 22                   & 93                   \\
\multicolumn{1}{|c|}{\textbf{Screw}}        & \multicolumn{1}{c|}{5}            & Object        & 320                        & 41                   & 119                  \\
\multicolumn{1}{|c|}{\textbf{Toothbrush}}   & \multicolumn{1}{c|}{1}            & Object        & 60                         & 12                   & 30                   \\ \hline
\multicolumn{1}{|c|}{\textbf{MVTec AD}}       & \multicolumn{1}{c|}{73}           & -             & 3,629                      & 467                  & 1,258                \\
\multicolumn{1}{|c|}{\textbf{AITEX}}        & \multicolumn{1}{c|}{12}           & Texture       & 1,692                      & 564                  & 183                  \\
\multicolumn{1}{|c|}{\textbf{SDD}}          & \multicolumn{1}{c|}{1}            & Texture       & 594                        & 286                  & 54                   \\
\multicolumn{1}{|c|}{\textbf{ELPV}}         & \multicolumn{1}{c|}{2}            & Texture       & 1,131                      & 377                  & 715                  \\
\multicolumn{1}{|c|}{\textbf{Optical}}      & \multicolumn{1}{c|}{1}            & Object        & 10,500                     & 3,500                & 2,100                \\
\multicolumn{1}{|c|}{\textbf{Mastcam}}      & \multicolumn{1}{c|}{11}           & Object        & 9,302                      & 426                  & 451                  \\
\multicolumn{1}{|c|}{\textbf{BrainMRI}}     & \multicolumn{1}{c|}{1}            & Medical       & 73                         & 25                   & 155                  \\
\multicolumn{1}{|c|}{\textbf{HeadCT}}       & \multicolumn{1}{c|}{1}            & Medical       & 75                         & 25                   & 100                  \\
\multicolumn{1}{|c|}{\textbf{Hyper-Kvasir}} & \multicolumn{1}{c|}{4}            & Medical       & 2,021                      & 674                  & 757                  \\ \hline
\end{tabular}}
\caption{Statistical details for nine real-world AD datasets, with the first 15 rows displaying detailed information for subsets of the MVTec AD dataset.}
\label{datset detail}
\end{table}

\vspace{0.1cm}
\noindent\textbf{MVTec AD}~\cite{bergmann2019mvtec} is a widely-used dataset that enables researchers to benchmark the performance of anomaly detection methods in the context of industrial inspection applications. The dataset includes over 5,000 images that are divided into 15 object and texture categories. Each category contains a training set of anomaly-free images, as well as a test set that includes images with both defects and defect-free images. 

\vspace{0.1cm}
\noindent\textbf{AITEX}~\cite{silvestre2019public} is a textile fabric database that comprises 245 images of 7 different fabrics, including 140 defect-free images (20 for each type of fabric) and 105 images with various types of defects. 

\vspace{0.1cm}
\noindent\textbf{SDD}~\cite{tabernik2020segmentation} is a collection of images captured in a controlled industrial environment, using defective production items as the subject. The dataset includes 52 images with visible defects and 347 product images without any defects. 

\vspace{0.1cm}
\noindent\textbf{ELPV}~\cite{deitsch2019automatic} is a collection of 2,624 high-resolution grayscale images of solar cells extracted from photovoltaic modules. These images were extracted from 44 different solar modules, and include both intrinsic and extrinsic defects known to reduce the power efficiency of solar modules. 

\vspace{0.1cm}
\noindent\textbf{Optical}~\cite{wieler2007weakly} is a synthetic dataset created to simulate real-world industrial inspection tasks for defect detection. The dataset comprises ten individual subsets, with the first six subsets (referred to as development datasets) intended for algorithm development purposes. The remaining four subsets (known as competition datasets) can be used to evaluate algorithm performance. 

\vspace{0.1cm}
\noindent\textbf{Mastcam}~\cite{kerner2020comparison} is a novelty detection dataset constructed from geological images captured by a multispectral imaging system installed on Mars exploration rovers. The dataset comprises typical images as well as images of 11 novel geologic classes. Each image includes a shorter wavelength (color) channel and a longer wavelengths (grayscale) channel. 

\vspace{0.1cm}
\noindent\textbf{BrainMRI}~\cite{salehi2021multiresolution} is a dataset for brain tumor detection obtained from magnetic resonance imaging (MRI) of the brain. 

\vspace{0.1cm}
\noindent\textbf{HeadCT}~\cite{salehi2021multiresolution} is a dataset consisting of 100 normal head CT slices and 100 slices with brain hemorrhage, without distinction between the types of hemorrhage. Each slice is from a different person, providing a diverse set of images for researchers to develop and test algorithms for hemorrhage detection and classification in medical imaging applications. 

\vspace{0.1cm}
\noindent\textbf{Hyper-Kvasir}~\cite{borgli2020hyperkvasir} is a large-scale open gastrointestinal dataset which is collected during real gastro- and colonoscopy procedures. It is comprised of four distinct parts, including labeled image data, unlabeled image data, segmented image data, and annotated video data.

\section{Implementation Details}
\subsection{Generating Partial Anomaly Distribution Datasets}\label{subsec:partial}
Our proposed approach creates a diverse collection of data subsets by randomly selecting a subset of normal clusters and incorporating labeled anomaly examples to form the support and query sets for learning partial anomaly distributions. Specifically, we generate each data subset as follows.

\vspace{0.1cm}
\noindent\textbf{Normal Samples in Each Data Subset.}
We employ the K-means algorithm to cluster normal samples into three groups.
In each data subset, we randomly select two clusters to form the support and query sets. We follow this way to generate six such data subsets. Furthermore, we include an additional subset consisting of all normal samples to capture a holistic view of the distribution of normal instances,
in which normal samples are randomly divided into two parts to form the support and query sets.

\begin{algorithm}
	\renewcommand{\algorithmicrequire}{\textbf{Input:}}
	\renewcommand{\algorithmicensure}{\textbf{Output:}}
	\caption{Anomaly Heterogeneity Learning (\coolname)}
	\label{alg2}
	\begin{algorithmic}[1]
        \REQUIRE Input $\mathcal{D}=\{\mathbf{x}, y\}, \{\phi\}_1^T, \psi$
        \ENSURE Output $g$
        \STATE /* Heterogeneous Anomaly Distribution Generation */
	\STATE Construct $\mathcal{D}_i$ through grouping training set $\mathcal{D}$ into $T$ groups
        \STATE /* Collaborative Differentiable Learning */
	\FOR {$epoch = 1$ to $N$}
		\STATE Update parameter of base model $\phi_i$ for $\mathcal{D}_i$ based on Eq.(1)
            \STATE /* Learning Importance Scores of Individual Anomaly Distributions */
		\IF {$epoch >= 5$}
            \STATE Compute the generalization error $r_i$ and importance score $w_i$ for $\phi_i$ with the help of sequential model $\psi$ via Eq.(6) and Eq.(7), respectively
            \ELSE
            \STATE Treat all $\phi$ equally, $w_i$ = $\frac{1}{T}$
            \ENDIF
            \STATE Update parameter of $g$ based on Eq.(4)
            \STATE Set the parameters of $\phi_i$ as the new weight parameters of $g$
            \ENDFOR
        \end{algorithmic}  
\end{algorithm}

\vspace{0.1cm}
\noindent\textbf{Abnormal Samples in Each Data Subset.}
To simulate diverse open-set environments in partial anomaly distribution subsets, for the setting with $M=10$, we choose $50\%$ of the seen anomalies randomly as virtual seen anomalies, which are present in both the support and query sets. The remaining $50\%$ are virtual unseen anomalies that are only available in the query set. In the case of $M=1$, where we only have one seen anomaly instance, both the training and refining sets have access to this sample. 

To enhance the variety of anomaly samples in our approach, we introduce three distinct anomaly augmentation techniques to generate pseudo anomaly samples: CutMix~\cite{yun2019cutmix}, CutPaste~\cite{li2021cutpaste}, DREAM Mask~\cite{zavrtanik2021draem}. These techniques are randomly applied to each partial anomaly distribution learning data subset to introduce diverse types of anomalies. 

\subsection{The Algorithm of \coolname}
The overall objective of our \coolname framework is to achieve a unified and robust AD model via synthesizing anomaly heterogeneities learned from various heterogeneous anomaly distributions. We summarize the Anomaly Heterogeneity Learning (\coolname) procedure in Algorithm~\ref{alg2}. Specifically, our framework first generates $T$ heterogeneous anomaly datasets, with each subset is sampled from the training set and contains a mixture of normal samples and (pseudo) anomaly samples, denoted as $\{\mathcal{D}_i\}_{i=1}^T$. In doing so, each subset is characterized by different set of normality/abnormality patterns,embodying heterogeneous anomaly distributions. We then employ a set of base models, denoted as $\{\phi_i\}_{i=1}^{T}$, to learn the underlying anomalous heterogeneity from these heterogeneous anomaly distributions. Moreover, a self-supervised sequential modeling approach is introduced to estimate the generalization errors $r_i$ and importance scores $w_i$ for each base model. Finally, we incorporate knowledge learned from heterogeneous anomaly distributions into a unified heterogeneous abnormality detection model $g$ to capture richer anomaly heterogeneity. 

\coolname is a generic framework, in which off-the-shelf open-set anomaly detectors can be easily plugged and gain significantly improved generalization and accuracy in detecting both seen and unseen anomalies. Once we choose a base anomaly detector, the training strategy and objective function should be consistent with it. Following the proposed loss of base models (\ie, DRA and DevNet), we adopt the deviation loss~\cite{pang2021explainable} to evaluate the loss between predicted anomaly scores and ground truths in the whole training phase:
\begin{align}
    \ell_{dev}(\mathbf{x},y;h) = &\mathbb{I}(y=0)|(dev(\mathbf{x};h))| \notag\\
    &+ \mathbb{I}(y=1)\max(0, m - dev(\mathbf{x};h)),  \notag
    \label{eq:loss_dev_loss}
\end{align}
where $\mathbb{I}(.)$ is an indicator function that is equal to one when the condition is true, and zero otherwise; $h(\cdot)$ denotes the anomaly detection model. $dev(\mathbf{x}) = \frac{h(\mathbf{x}) - \mu_r}{\sigma_r}$ with $\mu_r$ and $\sigma_r$ representing the mean and standard deviation of a set of sampled anomaly scores from the Gaussian prior distribution $\mathcal{N}(0, 1)$. $m$ is a confidence margin which defines a radius around the deviation.

\section{Detailed Empirical Results}
\subsection{Full Results under General Setting}
Table~\ref{gen_detail} shows the detailed comparison results of \coolname and SOTA competing methods under the general setting. It includes the performance metrics of each category of MVTec AD dataset. Overall, our proposed \coolname model consistently outperforms the baseline methods in both ten-shot and one-shot settings across all three application scenarios. \coolname (DRA) achieves the best performance in terms of AUC. On average, \coolname improves the AUC of DRA and DevNet by up to $4\%$ and $9\%$, respectively. 

\begin{table*}[ht]
\resizebox{\linewidth}{!}{
\begin{tabular}{c|ccccccc|ccccccc}
\hline
& \multicolumn{7}{c|}{\textbf{One Anomaly Examples (Random)}}                                                                                                                                                                                                                             & \multicolumn{7}{c}{\textbf{Ten Anomaly Examples (Random)}}                                                                                                                             \\ \cline{2-15} 
\multirow{-2}{*}{\textbf{Dataset}} & \textbf{SAOE} & \textbf{MLEP} & \textbf{FLOS} & \textbf{DevNet}      & \textbf{DRA}                                 & \textbf{AHL(DevNet)}                        & \textbf{AHL(DRA)}                                                   & \textbf{SAOE} & \textbf{MLEP} & \textbf{FLOS} & \textbf{DevNet}      & \textbf{DRA}         & \textbf{AHL(DevNet)}                        & \textbf{AHL(DRA)}                           \\ \hline
\textbf{Carpet}                                            & 0.766{\scriptsize±0.098}   & 0.701{\scriptsize±0.091}   & 0.755{\scriptsize±0.026}   & 0.778{\scriptsize±0.055}          & \textbf{0.873{\scriptsize±0.035}}                         & 0.802{\scriptsize±0.018}                                 & {\color[HTML]{FF0000} \textbf{0.877{\scriptsize±0.004}}}                         & 0.755{\scriptsize±0.136}   & 0.781{\scriptsize±0.049}   & 0.780{\scriptsize±0.009}   & 0.864{\scriptsize±0.012}          & \textbf{0.945{\scriptsize±0.014}}          & 0.867{\scriptsize±0.006}                                 & {\color[HTML]{FF0000} \textbf{0.953{\scriptsize±0.001}}}                                 \\
\textbf{Grid}                                              & 0.921{\scriptsize±0.032}   & 0.839{\scriptsize±0.028}   & 0.871{\scriptsize±0.076}   & 0.868{\scriptsize±0.031}          & \textbf{0.972{\scriptsize±0.016}} & 0.872{\scriptsize±0.032}                                 & {\color[HTML]{FF0000} \textbf{0.975{\scriptsize±0.005}}}                         & 0.952{\scriptsize±0.011}   & 0.980{\scriptsize±0.009}   & 0.966{\scriptsize±0.005}   & 0.901{\scriptsize±0.016 }         & \textbf{0.990{\scriptsize±0.008} }         & 0.914{\scriptsize±0.003}                                 & {\color[HTML]{FF0000} \textbf{0.992{\scriptsize±0.002}}}                                 \\
\textbf{Leather}                                           & 0.996{\scriptsize±0.007}   & 0.781{\scriptsize±0.020}   & 0.791{\scriptsize±0.057}   & 0.874{\scriptsize±0.016}          & {\color[HTML]{FF0000} \textbf{0.988{\scriptsize±0.003}}}  & \textbf{0.880{\scriptsize±0.005}}                        & {\color[HTML]{FF0000} \textbf{0.988{\scriptsize±0.001}} }                        & 1.000{\scriptsize±0.000}   & 0.813{\scriptsize±0.158}   & 0.993{\scriptsize±0.004}   & 0.986{\scriptsize±0.033 }         & {\color[HTML]{FF0000} \textbf{1.000{\scriptsize±0.000}}}          & \textbf{0.996{\scriptsize±0.008} }                                & {\color[HTML]{FF0000} \textbf{1.000{\scriptsize±0.000 }}}                                \\
\textbf{Tile}                                              & 0.935{\scriptsize±0.034}   & 0.927{\scriptsize±0.036}   & 0.787{\scriptsize±0.038}   & 0.872{\scriptsize±0.035}          & \textbf{0.966{\scriptsize±0.014}}                         & 0.909{\scriptsize±0.007 }                                & {\color[HTML]{FF0000} \textbf{0.968{\scriptsize±0.001}} }                        & 0.944{\scriptsize±0.013}   & \textbf{0.988{\scriptsize±0.009}}   & 0.952{\scriptsize±0.010}   & {\color[HTML]{FF0000} \textbf{1.000{\scriptsize±0.000}}}          & {\color[HTML]{FF0000} \textbf{1.000{\scriptsize±0.000}}}          & {\color[HTML]{FF0000} \textbf{1.000{\scriptsize±0.000}}}                                 & {\color[HTML]{FF0000} \textbf{1.000{\scriptsize±0.000}} }                                \\
\textbf{Wood}                                              & 0.948{\scriptsize±0.009}   & 0.660{\scriptsize±0.142}   & 0.927{\scriptsize±0.065}   & 0.917{\scriptsize±0.029 }         & {\color[HTML]{FF0000} \textbf{0.987{\scriptsize±0.012}}}  & \textbf{0.947{\scriptsize±0.020}}                        & {\color[HTML]{FF0000} \textbf{0.987{\scriptsize±0.003}} }                        & 0.976{\scriptsize±0.031}   & \textbf{0.999{\scriptsize±0.002}}   & {\color[HTML]{FF0000} \textbf{1.000{\scriptsize±0.000}}}   & \textbf{0.999{\scriptsize±0.000} }         & 0.998{\scriptsize±0.013 }         & {\color[HTML]{FF0000} \textbf{1.000{\scriptsize±0.000 }}}                                & 0.998{\scriptsize±0.000}                                \\
\textbf{Bottle}                                            & 0.989{\scriptsize±0.019}   & 0.927{\scriptsize±0.090}   & 0.975{\scriptsize±0.023}   & 0.986{\scriptsize±0.012}          & {\color[HTML]{FF0000} \textbf{1.000{\scriptsize±0.000}}}  & \textbf{0.994{\scriptsize±0.004}}                        & {\color[HTML]{FF0000} \textbf{1.000{\scriptsize±0.000}}}                         & 0.998{\scriptsize±0.003}   & 0.981{\scriptsize±0.004}   & 0.995{\scriptsize±0.002}   & 0.996{\scriptsize±0.005}          & {\color[HTML]{FF0000} \textbf{1.000{\scriptsize±0.000}}}          & \textbf{0.998{\scriptsize±0.001}}                                 & {\color[HTML]{FF0000} \textbf{1.000{\scriptsize±0.000}}}                                 \\
\textbf{Capsule}                                           & 0.611{\scriptsize±0.109}   & 0.558{\scriptsize±0.075}   & 0.666{\scriptsize±0.020}   & 0.567{\scriptsize±0.042}          & \textbf{0.646{\scriptsize±0.029}}                         & 0.581{\scriptsize±0.202}                                 & {\color[HTML]{FF0000} \textbf{0.665{\scriptsize±0.030}}}                         & 0.850{\scriptsize±0.054}   & 0.818{\scriptsize±0.063}   & 0.902{\scriptsize±0.017}   & 0.872{\scriptsize±0.017}          & \textbf{0.928{\scriptsize±0.011 }}         & 0.885{\scriptsize±0.012}                                 & {\color[HTML]{FF0000} \textbf{0.930{\scriptsize±0.001 }}}                                \\
\textbf{Pill}                                              & 0.652{\scriptsize±0.078}   & 0.656{\scriptsize±0.061}   & 0.745{\scriptsize±0.064}   & 0.779{\scriptsize±0.018}          & \textbf{0.831{\scriptsize±0.026}}                         & 0.781{\scriptsize±0.087}                                 & {\color[HTML]{FF0000} \textbf{0.840{\scriptsize±0.003}}}                         & 0.872{\scriptsize±0.049}   & 0.845{\scriptsize±0.048}   & {\color[HTML]{FF0000} \textbf{0.929{\scriptsize±0.012}}}   & 0.882{\scriptsize±0.008}          & \textbf{0.918{\scriptsize±0.009} }         & 0.900{\scriptsize±0.004}                                 & \textbf{0.918{\scriptsize±0.001 }}                               \\
\textbf{Transistor}                                        & 0.680{\scriptsize±0.182}   & 0.695{\scriptsize±0.124 }  & 0.709{\scriptsize±0.041 }  & 0.732{\scriptsize±0.075}          & 0.727{\scriptsize±0.105}                                  & \textbf{0.737{\scriptsize±0.098} }                       & {\color[HTML]{FF0000} \textbf{0.796{\scriptsize±0.003}} }                        & 0.860{\scriptsize±0.053}   & 0.927{\scriptsize±0.043}   & 0.862{\scriptsize±0.037}   & 0.907{\scriptsize±0.004 }         & \textbf{0.919{\scriptsize±0.003}}          & 0.912{\scriptsize±0.002 }                                & {\color[HTML]{FF0000} \textbf{0.926{\scriptsize±0.009}} }                                \\
\textbf{Zipper}                                            & 0.970{\scriptsize±0.033}   & 0.856{\scriptsize±0.086}   & 0.885{\scriptsize±0.033}   & 0.914{\scriptsize±0.027}          & \textbf{0.983{\scriptsize±0.008}} & 0.928{\scriptsize±0.006}                                 & {\color[HTML]{FF0000} \textbf{0.986{\scriptsize±0.000}}}                         & 0.995{\scriptsize±0.004}   & 0.965{\scriptsize±0.002}   & 0.990{\scriptsize±0.008}   & 0.992{\scriptsize±0.008}          & {\color[HTML]{FF0000} \textbf{1.000{\scriptsize±0.000}}}          & \textbf{0.995{\scriptsize±0.002}}                                 & {\color[HTML]{FF0000} \textbf{1.000{\scriptsize±0.000} }}                                \\
\textbf{Cable}                                             & 0.819{\scriptsize±0.060}   & 0.688{\scriptsize±0.017}   & 0.790{\scriptsize±0.039 }  & 0.790{\scriptsize±0.086 }         & \textbf{0.855{\scriptsize±0.007}} & 0.793{\scriptsize±0.091}                                 & {\color[HTML]{FF0000} \textbf{0.858{\scriptsize±0.011}}} & 0.862±0.022   & 0.857{\scriptsize±0.062}   & 0.890{\scriptsize±0.063}   & 0.901{\scriptsize±0.006 }         & \textbf{0.914{\scriptsize±0.006}}          & 0.907{\scriptsize±0.004}                                 & {\color[HTML]{FF0000} \textbf{0.921{\scriptsize±0.001}}}                                 \\
\textbf{Hazelnut}                                          & 0.961{\scriptsize±0.042}   & 0.704{\scriptsize±0.090}   & 0.976{\scriptsize±0.021}   & 0.970{\scriptsize±0.005}          & \textbf{0.982{\scriptsize±0.005}} & 0.978{\scriptsize±0.003}                                 & {\color[HTML]{FF0000} \textbf{0.989{\scriptsize±0.004}}}                         & {\color[HTML]{FF0000} \textbf{1.000{\scriptsize±0.000}}}   & {\color[HTML]{FF0000} \textbf{1.000{\scriptsize±0.000}}}   & {\color[HTML]{FF0000} \textbf{1.000{\scriptsize±0.000}}}   & {\color[HTML]{FF0000} \textbf{1.000{\scriptsize±0.000}}}          & {\color[HTML]{FF0000} \textbf{1.000{\scriptsize±0.000}}}          & {\color[HTML]{FF0000} \textbf{1.000{\scriptsize±0.000}}}                                 & {\color[HTML]{FF0000} \textbf{1.000{\scriptsize±0.000}}}                                \\
\textbf{Metal\_nut}                                        & 0.922{\scriptsize±0.033}   & 0.878{\scriptsize±0.038}   & 0.930{\scriptsize±0.022}   & 0.874{\scriptsize±0.014}          & \textbf{0.950{\scriptsize±0.011}}                         & 0.880{\scriptsize±0.002}                                 & {\color[HTML]{FF0000} \textbf{0.952{\scriptsize±0.003}}}                         & 0.976{\scriptsize±0.013}   & 0.974{\scriptsize±0.009}   & 0.984{\scriptsize±0.004}   & 0.992{\scriptsize±0.003}          & \textbf{0.997{\scriptsize±0.005}}          & \textbf{0.997{\scriptsize±0.003}}                                 & {\color[HTML]{FF0000} \textbf{0.998{\scriptsize±0.000}} }                                \\
\textbf{Screw}                                             & 0.653{\scriptsize±0.074}   & 0.675{\scriptsize±0.294}   & 0.337{\scriptsize±0.091}   & 0.766{\scriptsize±0.045}          & \textbf{0.902{\scriptsize±0.032}}                         & 0.769{\scriptsize±0.031}                                 & {\color[HTML]{FF0000} \textbf{0.927{\scriptsize±0.009}}} & 0.975{\scriptsize±0.023}   & 0.899{\scriptsize±0.039}   & 0.940{\scriptsize±0.017 }  & 0.981{\scriptsize±0.013}          & 0.978{\scriptsize±0.012}          & \textbf{0.984{\scriptsize±0.005}}                                 & {\color[HTML]{FF0000} \textbf{0.985{\scriptsize±0.002}}}                                 \\
\textbf{Toothbrush}                                        & 0.686{\scriptsize±0.110}   & 0.617{\scriptsize±0.058}   & 0.731{\scriptsize±0.028}   & \textbf{0.790{\scriptsize±0.029}} & 0.675{\scriptsize±0.019}                                  & {\color[HTML]{FF0000} \textbf{0.794{\scriptsize±0.016}}} & 0.710{\scriptsize±0.007 }                                                        & 0.865{\scriptsize±0.062}   & 0.783{\scriptsize±0.048}   & 0.900{\scriptsize±0.008}   & \textbf{0.950{\scriptsize±0.025}}          & 0.908{\scriptsize±0.007 }         & {\color[HTML]{FF0000} \textbf{0.959{\scriptsize±0.002}}}                                 & 0.921{\scriptsize±0.007}                                 \\ \hline
\textbf{MVTec AD}                                      & 0.834{\scriptsize±0.007}   & 0.744{\scriptsize±0.019}   & 0.792{\scriptsize±0.014}   & 0.832{\scriptsize±0.016 }         & \textbf{0.889{\scriptsize±0.013}}                         & 0.843{\scriptsize±0.021}                                 & {\color[HTML]{FF0000} \textbf{0.901{\scriptsize±0.003}}}                         & 0.926{\scriptsize±0.010}   & 0.907{\scriptsize±0.005}   & 0.939{\scriptsize±0.007}   & 0.948{\scriptsize±0.005} & \textbf{0.966{\scriptsize±0.002}} & 0.954{\scriptsize±0.003} & {\color[HTML]{FF0000} \textbf{0.970{\scriptsize±0.002}}} \\
\textbf{AITEX}                                             & 0.675{\scriptsize±0.094}   & 0.564{\scriptsize±0.055}   & 0.538{\scriptsize±0.073}   & 0.609{\scriptsize±0.054}          & 0.693{\scriptsize±0.031}                                  & \textbf{0.704{\scriptsize±0.004}} & {\color[HTML]{FF0000} \textbf{0.734{\scriptsize±0.008}}}                                                & 0.874{\scriptsize±0.024}   & 0.867{\scriptsize±0.037}   & 0.841{\scriptsize±0.049}   & 0.889{\scriptsize±0.007} & 0.892{\scriptsize±0.007} & \textbf{0.903{\scriptsize±0.011}} & {\color[HTML]{FF0000} \textbf{0.925{\scriptsize±0.013}}} \\
\textbf{SDD}                                               & 0.781{\scriptsize±0.009}   & 0.811{\scriptsize±0.045 }  & 0.840{\scriptsize±0.043}   & 0.851{\scriptsize±0.003 }         & \textbf{0.907{\scriptsize±0.002} }                        & 0.864{\scriptsize±0.001}          & {\color[HTML]{FF0000} \textbf{0.909{\scriptsize±0.001}} }                                               & 0.955{\scriptsize±0.020}   & 0.783{\scriptsize±0.013 }  & 0.967{\scriptsize±0.018}   & 0.985{\scriptsize±0.004} & \textbf{0.990{\scriptsize±0.000}} & {\color[HTML]{FF0000} \textbf{0.991{\scriptsize±0.001}}} & {\color[HTML]{FF0000} \textbf{0.991{\scriptsize±0.000}} }\\
\textbf{ELPV}                                              & 0.635{\scriptsize±0.092}   & 0.578{\scriptsize±0.062}   & 0.457{\scriptsize±0.056}   & \textbf{0.810{\scriptsize±0.024}} & 0.676{\scriptsize±0.003 }                                 & {\color[HTML]{FF0000} \textbf{0.828{\scriptsize±0.005}}} & {\color[HTML]{333333} 0.723{\scriptsize±0.008}}                                  & 0.793{\scriptsize±0.047 }  & 0.794{\scriptsize±0.047}   & 0.818{\scriptsize±0.032}   & 0.843{\scriptsize±0.001} & 0.843{\scriptsize±0.002} & \textbf{0.849{\scriptsize±0.003} }& {\color[HTML]{FF0000} \textbf{0.850{\scriptsize±0.004}}} \\
\textbf{Optical}                                           & 0.815{\scriptsize±0.014}   & 0.516{\scriptsize±0.009 }  & 0.518{\scriptsize±0.003}   & 0.513{\scriptsize±0.001}          & \textbf{0.880{\scriptsize±0.002}}                         & 0.547{\scriptsize±0.009}          & {\color[HTML]{FF0000} \textbf{0.888{\scriptsize±0.007}}}                                                & 0.941{\scriptsize±0.013}   & 0.740{\scriptsize±0.039}   & 0.720{\scriptsize±0.055}   & 0.785{\scriptsize±0.012 }& \textbf{0.966{\scriptsize±0.002}} & 0.841{\scriptsize±0.010} & {\color[HTML]{FF0000} \textbf{0.976{\scriptsize±0.004}}} \\
\textbf{Mastcam}                                           & 0.662{\scriptsize±0.018}   & 0.625{\scriptsize±0.045}   & 0.542{\scriptsize±0.017 }  & 0.627{\scriptsize±0.049 }         & \textbf{0.709{\scriptsize±0.011} }                        & {\color[HTML]{333333} 0.644{\scriptsize±0.013}}          & {\color[HTML]{FF0000} \textbf{0.743{\scriptsize±0.003} } }                                              & 0.810{\scriptsize±0.029}   & 0.798{\scriptsize±0.026}   & 0.703{\scriptsize±0.029}   & 0.797{\scriptsize±0.021} & \textbf{0.849{\scriptsize±0.003}} & 0.825{\scriptsize±0.020} & {\color[HTML]{FF0000} \textbf{0.855{\scriptsize±0.005}}} \\
\textbf{BrainMRI}                                          & 0.531{\scriptsize±0.060}   & 0.632{\scriptsize±0.017 }  & 0.693{\scriptsize±0.036}   & \textbf{0.853{\scriptsize±0.045}} & 0.747{\scriptsize±0.001}                                  & {\color[HTML]{FF0000} \textbf{0.866{\scriptsize±0.004}}} & 0.760{\scriptsize±0.013 }                            & 0.900{\scriptsize±0.041}   & 0.959{\scriptsize±0.011 }  & 0.955{\scriptsize±0.011}   & 0.951{\scriptsize±0.007} & \textbf{0.971{\scriptsize±0.001}} & 0.959{\scriptsize±0.008} & {\color[HTML]{FF0000} \textbf{0.977{\scriptsize±0.001}}} \\
\textbf{HeadCT}                                            & 0.597{\scriptsize±0.022}   & 0.758{\scriptsize±0.038}   & 0.698{\scriptsize±0.092}   & 0.755{\scriptsize±0.029}          & \textbf{0.804{\scriptsize±0.010} }                        & {\color[HTML]{333333} 0.781{\scriptsize±0.007}}          & {\color[HTML]{FF0000} \textbf{0.825{\scriptsize±0.014} } }                                              & 0.935{\scriptsize±0.021}   & 0.972{\scriptsize±0.014 }  & 0.971{\scriptsize±0.004 }  & \textbf{0.997{\scriptsize±0.002}} & 0.988{\scriptsize±0.001} & {\color[HTML]{FF0000} \textbf{0.999{\scriptsize±0.003}}} & 0.993{\scriptsize±0.002} \\
\textbf{Hyper-Kvasir}                                      & 0.498{\scriptsize±0.100}   & 0.445{\scriptsize±0.040}   & 0.668{\scriptsize±0.004}   & 0.734{\scriptsize±0.020}          & 0.712{\scriptsize±0.010 }                                 & {\color[HTML]{FF0000} \textbf{0.768{\scriptsize±0.015}}} & {\color[HTML]{333333} \textbf{0.742{\scriptsize±0.015}}}                         & 0.666{\scriptsize±0.050}   & 0.600{\scriptsize±0.069 }  & 0.773{\scriptsize±0.029}   & 0.822{\scriptsize±0.031} & 0.844{\scriptsize±0.001}&  \textbf{0.873{\scriptsize±0.009}} & {\color[HTML]{FF0000} \textbf{0.880{\scriptsize±0.003}}} \\ \hline
\end{tabular}}
\caption{AUC results(mean±std) on nine real-world AD datasets under the general setting. Best results and the second-best results are respectively highlighted in {\color[HTML]{FF0000} \textbf{Red}} and \textbf{Bold}.}
\label{gen_detail}
\end{table*}

\subsection{Full Results under Hard Setting}
To investigate the detection performance of \coolname framework on novel anomaly classes, we evaluate its performance under the hard setting, and present the detailed results for six multi-subset datasets, including each anomaly class-level performance, in Table~\ref{hard_detail}. Overall, our models -- \coolname (DRA) and \coolname (DevNet) -- achieve the best AUC results in both $M=1$ and $M=10$ setting protocols. Specifically, \coolname improves the performances of DRA and DevNet by up to $3.2\%$ and $3\%$ AUC, respectively. The results here are consistent with the superiority performance of \coolname in the general setting.

\begin{table*}[ht]
\resizebox{\linewidth}{!}{
\begin{tabular}{cc|lllclcl|lllclcc}
\hline
\multicolumn{2}{c|}{\multirow{2}{*}{\textbf{Dataset}}}                                    & \multicolumn{7}{c|}{\textbf{One Example from One Anomaly Class}}                                                                  & \multicolumn{7}{c}{\textbf{Ten Example from One Anomaly Class}}                                                                                                                                      \\ \cline{3-16} 
\multicolumn{2}{c|}{}                                                                     & \textbf{SAOE}         & \textbf{MLEP}        & \textbf{FLOS}        & \multicolumn{1}{c}{\textbf{DevNet}} & \textbf{DRA} & \multicolumn{1}{c}{\textbf{AHL(DevNet)}} & \textbf{AHL(DRA)} & \textbf{SAOE}                            & \textbf{MLEP}        & \textbf{FLOS}        & \multicolumn{1}{c}{\textbf{DevNet}} & \textbf{DRA}                             & \multicolumn{1}{c}{\textbf{AHL(DevNet)}} & \multicolumn{1}{c}{\textbf{AHL(DRA)}} \\ \hline
\multicolumn{1}{c|}{\multirow{6}{*}{\textbf{Carpet}}}       & \textbf{Color}              & 0.763{\scriptsize±0.100}  & 0.547{\scriptsize±0.056} & 0.467{\scriptsize±0.278} & 0.701{\scriptsize±0.046}                &   \textbf{0.890{\scriptsize±0.011}}  & 0.718{\scriptsize±0.009}                     &    {\color[HTML]{FF0000} \textbf{0.894{\scriptsize±0.004}}}      & 0.467{\scriptsize±0.067}                     & 0.698{\scriptsize±0.025} & 0.760{\scriptsize±0.005} & 0.774{\scriptsize±0.009}                & \textbf{0.899{\scriptsize±0.019}}                     & 0.778{\scriptsize±0.004}                     & {\color[HTML]{FF0000} \textbf{0.929{\scriptsize±0.007}} }                 \\
\multicolumn{1}{c|}{}                                       & \textbf{Cut}                & 0.664{\scriptsize±0.165}  & 0.658{\scriptsize±0.056} & 0.685{\scriptsize±0.007} & 0.679{\scriptsize±0.018 }               &   \textbf{0.890{\scriptsize±0.024}}  & 0.684{\scriptsize±0.014}                     &    {\color[HTML]{FF0000} \textbf{0.934{\scriptsize±0.003}}}      & 0.793{\scriptsize±0.175}                     & 0.653{\scriptsize±0.120} & 0.688{\scriptsize±0.059} & 0.817{\scriptsize±0.021}                & \textbf{0.942{\scriptsize±0.012}}                     & 0.825{\scriptsize±0.006 }                    & {\color[HTML]{FF0000} \textbf{0.943{\scriptsize±0.002}}}                  \\
\multicolumn{1}{c|}{}                                       & \textbf{Hole}               & 0.772{\scriptsize±0.071}  & 0.653{\scriptsize±0.065} & 0.594{\scriptsize±0.142} & 0.729{\scriptsize±0.032}                &   \textbf{0.915{\scriptsize±0.045}}  & 0.736{\scriptsize±0.062 }                    &     {\color[HTML]{FF0000} \textbf{0.935{\scriptsize±0.014}} }    & 0.831{\scriptsize±0.125}                     & 0.674{\scriptsize±0.076} & 0.733{\scriptsize±0.014} & 0.808{\scriptsize±0.016}                & \textbf{0.958{\scriptsize±0.031}}                     & 0.815{\scriptsize±0.036}                     & {\color[HTML]{FF0000} \textbf{0.960{\scriptsize±0.003}}}                  \\
\multicolumn{1}{c|}{}                                       & \textbf{Metal}              & 0.780{\scriptsize±0.172}  & 0.706{\scriptsize±0.047} & 0.701{\scriptsize±0.028} & 0.822{\scriptsize±0.016 }               &   \textbf{0.877{\scriptsize±0.013}}  & 0.846{\scriptsize±0.012}                    &    {\color[HTML]{FF0000} \textbf{0.931{\scriptsize±0.007}}}      & \multicolumn{1}{c}{0.883{\scriptsize±0.043}} & 0.764{\scriptsize±0.061} & 0.678{\scriptsize±0.083} & 0.885{\scriptsize±0.012}                & \multicolumn{1}{c}{\textbf{0.916{\scriptsize±0.017}}} & 0.899{\scriptsize±0.026 }                    & {\color[HTML]{FF0000} \textbf{0.921{\scriptsize±0.003}}}                  \\
\multicolumn{1}{c|}{}                                       & \textbf{Thread}             & 0.787{\scriptsize±0.204}  & 0.831{\scriptsize±0.117} & 0.941{\scriptsize±0.005} & 0.937{\scriptsize±0.017}                &  \textbf{0.954{\scriptsize±0.010}}   & 0.941{\scriptsize±0.006}                     &     {\color[HTML]{FF0000} \textbf{0.966{\scriptsize±0.005}}}     & \multicolumn{1}{c}{0.834{\scriptsize±0.297}} & 0.967{\scriptsize±0.006} & 0.946{\scriptsize±0.005} & 0.981{\scriptsize±0.005}                & \multicolumn{1}{c}{\textbf{0.985{\scriptsize±0.005}}} & 0.984{\scriptsize±0.013}                     & {\color[HTML]{FF0000} \textbf{0.991{\scriptsize±0.001}}}                  \\ \cline{2-16} 
\multicolumn{1}{c|}{}                                       & \textbf{Mean}               & 0.753{\scriptsize±0.055}  & 0.679{\scriptsize±0.029} & 0.678{\scriptsize±0.040} & 0.774{\scriptsize±0.007}                &   \textbf{0.905{\scriptsize±0.006}}  & 0.785{\scriptsize±0.015}                     &     {\color[HTML]{FF0000} \textbf{0.932{\scriptsize±0.003}}}     & 0.762{\scriptsize±0.073}                     & 0.751{\scriptsize±0.023} & 0.761{\scriptsize±0.012} & 0.853{\scriptsize±0.005}                & \textbf{0.940{\scriptsize±0.006}}                     & 0.860{\scriptsize±0.013}                     & {\color[HTML]{FF0000} \textbf{0.949{\scriptsize±0.002 }}}                 \\ \hline
\multicolumn{1}{c|}{\multirow{5}{*}{\textbf{Metal\_nut}}}   & \textbf{Bent}               & 0.864{\scriptsize±0.032}  & 0.743{\scriptsize±0.013} & 0.851{\scriptsize±0.046} & 0.817{\scriptsize±0.033}                &   \textbf{0.952{\scriptsize±0.015}}  & 0.831{\scriptsize±0.020 }                    &     {\color[HTML]{FF0000} \textbf{0.954{\scriptsize±0.003}}}     & 0.901{\scriptsize±0.023}                     & 0.956{\scriptsize±0.013} & 0.827{\scriptsize±0.075} & 0.907{\scriptsize±0.018}                & \textbf{0.987{\scriptsize±0.003}}                     & 0.909{\scriptsize±0.016}                     & {\color[HTML]{FF0000} \textbf{0.989{\scriptsize±0.000}}}                  \\
\multicolumn{1}{c|}{}                                       & \textbf{Color}              & 0.857{\scriptsize±0.037}  & 0.835{\scriptsize±0.075} & 0.821{\scriptsize±0.059} & 0.903{\scriptsize±0.019}                &  \textbf{0.930{\scriptsize±0.021}}   & 0.910{\scriptsize±0.008}                     &    {\color[HTML]{FF0000} \textbf{0.933{\scriptsize±0.008}}}      & 0.879{\scriptsize±0.018}                     & 0.945{\scriptsize±0.039} & 0.978{\scriptsize±0.008} & \textbf{0.992{\scriptsize±0.015}}                & 0.956{\scriptsize±0.009}                     & {\color[HTML]{FF0000} \textbf{0.995{\scriptsize±0.002}}}                     & 0.958{\scriptsize±0.001}                  \\
\multicolumn{1}{c|}{}                                       & \textbf{Flip}               & 0.751{\scriptsize±0.090}  & 0.813{\scriptsize±0.031} & 0.799{\scriptsize±0.058} & 0.751{\scriptsize±0.039}                &  {\color[HTML]{FF0000} \textbf{0.931{\scriptsize±0.017}}}   & \textbf{0.755{\scriptsize±0.022}}                     &      {\color[HTML]{FF0000} \textbf{0.931{\scriptsize±0.001}}}    & 0.795{\scriptsize±0.062 }                    & 0.805{\scriptsize±0.057} & 0.942{\scriptsize±0.009} & \textbf{0.982{\scriptsize±0.010}}                & 0.931{\scriptsize±0.010}                     & {\color[HTML]{FF0000} \textbf{0.987{\scriptsize±0.003}}}                     & 0.937{\scriptsize±0.003}                  \\
\multicolumn{1}{c|}{}                                       & \textbf{Scratch}            & 0.792{\scriptsize±0.075}  & 0.907{\scriptsize±0.085} & 0.947{\scriptsize±0.027 }& \textbf{0.974{\scriptsize±0.061}}                &  0.929{\scriptsize±0.009}   & {\color[HTML]{FF0000} \textbf{0.981{\scriptsize±0.035}}}                     &   0.934{\scriptsize±0.005}       & 0.845{\scriptsize±0.041}                     & 0.805{\scriptsize±0.153} & 0.943{\scriptsize±0.002} & \textbf{0.998{\scriptsize±0.005}}                & \textbf{0.998{\scriptsize±0.006}}                    & \textbf{0.998{\scriptsize±0.001}}                     & {\color[HTML]{FF0000} \textbf{0.999{\scriptsize±0.00}}}                  \\ \cline{2-16} 
\multicolumn{1}{c|}{}                                       & \textbf{Mean}               & 0.816{\scriptsize±0.029}  & 0.825{\scriptsize±0.023} & 0.855{\scriptsize±0.024} & 0.861{\scriptsize±0.019}                &  \textbf{0.936{\scriptsize±0.011}}   & 0.869{\scriptsize±0.004 }                    &      {\color[HTML]{FF0000} \textbf{0.939{\scriptsize±0.004}} }   & 0.855{\scriptsize±0.016}                     & 0.878{\scriptsize±0.058} & 0.922{\scriptsize±0.014} & 0.970{\scriptsize±0.009}                & 0.968{\scriptsize±0.006}                     & {\color[HTML]{FF0000} \textbf{0.972{\scriptsize±0.002}}}                     & \textbf{0.971{\scriptsize±0.001}}                  \\ \hline
\multicolumn{1}{c|}{\multirow{7}{*}{\textbf{AITEX}}}        & \textbf{Broken end}         & {\color[HTML]{FF0000} \textbf{0.778{\scriptsize±0.068}}}  & 0.441{\scriptsize±0.111} & 0.645{\scriptsize±0.030} & 0.702{\scriptsize±0.037}                &  0.696{\scriptsize±0.057}   & \textbf{0.716{\scriptsize±0.014}}                     &    0.704{\scriptsize±0.005}      & 0.712{\scriptsize±0.068}                     & \textbf{0.732{\scriptsize±0.065}} & 0.585{\scriptsize±0.037} & 0.658{\scriptsize±0.062}                & 0.708{\scriptsize±0.062}                     & 0.688{\scriptsize±0.013}                     & {\color[HTML]{FF0000} \textbf{0.735{\scriptsize±0.010}} }                 \\
\multicolumn{1}{c|}{}                                       & \textbf{Broken pick}        & 0.644{\scriptsize±0.039}  & 0.476{\scriptsize±0.070} & 0.598{\scriptsize±0.023} & 0.567{\scriptsize±0.016}                &   \textbf{0.719{\scriptsize±0.004}}  & 0.575{\scriptsize±0.005}                     &     {\color[HTML]{FF0000} \textbf{0.727{\scriptsize±0.003}} }    & 0.629{\scriptsize±0.012}                     & 0.555{\scriptsize±0.027} & 0.548{\scriptsize±0.054} & 0.595{\scriptsize±0.017}                & \textbf{0.671{\scriptsize±0.034}}                     & 0.612{\scriptsize±0.005 }                    & {\color[HTML]{FF0000} \textbf{0.683{\scriptsize±0.002}}}                  \\
    \multicolumn{1}{c|}{}                                       & \textbf{Cut selvage}        & 0.681{\scriptsize±0.077}  & 0.434{\scriptsize±0.149} & 0.694{\scriptsize±0.036} & 0.674{\scriptsize±0.021}                &   \textbf{0.751{\scriptsize±0.006}}  & 0.680{\scriptsize±0.017}                     &     {\color[HTML]{FF0000} \textbf{0.753{\scriptsize±0.007}}}     & 0.770{\scriptsize±0.014}                     & 0.682{\scriptsize±0.025} & 0.745{\scriptsize±0.035} & 0.703{\scriptsize±0.062}                & \textbf{0.777{\scriptsize±0.021}}                     & 0.737{\scriptsize±0.012}                     & {\color[HTML]{FF0000} \textbf{0.781{\scriptsize±0.006}}}                  \\
\multicolumn{1}{c|}{}                                       & \textbf{Fuzzyball}          & {\color[HTML]{FF0000} \textbf{0.650{\scriptsize±0.064}}}  & 0.525{\scriptsize±0.157} & 0.525{\scriptsize±0.043} & 0.629{\scriptsize±0.103}                &   0.631{\scriptsize±0.018}  & 0.644{\scriptsize±0.031}                     &     \textbf{0.647{\scriptsize±0.007}}      & {\color[HTML]{FF0000} \textbf{0.842{\scriptsize±0.026}}}                     & 0.677{\scriptsize±0.223} & 0.550{\scriptsize±0.082} & 0.736{\scriptsize±0.101}                & 0.749{\scriptsize±0.033}                     & 0.755{\scriptsize±0.002}                     & \textbf{0.775{\scriptsize±0.024}}                  \\
\multicolumn{1}{c|}{}                                       & \textbf{Nep}                & 0.710{\scriptsize±0.044}  & 0.517{\scriptsize±0.059} & 0.734{\scriptsize±0.038} & \textbf{0.741{\scriptsize±0.011}}                &   0.685{\scriptsize±0.010 } & {\color[HTML]{FF0000} \textbf{0.754{\scriptsize±0.012}} }                    &    0.703{\scriptsize±0.005}      & 0.771{\scriptsize±0.032}                     & 0.740{\scriptsize±0.052} & 0.746{\scriptsize±0.060} & \textbf{0.806{\scriptsize±0.039}}                & 0.784{\scriptsize±0.025}                     & {\color[HTML]{FF0000} \textbf{0.836{\scriptsize±0.007}}}                     & 0.792{\scriptsize±0.007}                  \\
\multicolumn{1}{c|}{}                                       & \textbf{Weft crack}         & 0.582{\scriptsize±0.108}  & 0.400{\scriptsize±0.029} & 0.546{\scriptsize±0.114} & 0.561{\scriptsize±0.085}                &   \textbf{0.693{\scriptsize±0.002}}  & 0.588{\scriptsize±0.018 }                    &    {\color[HTML]{FF0000} \textbf{0.706{\scriptsize±0.009}} }     & 0.618{\scriptsize±0.172}                     & 0.370{\scriptsize±0.037} & 0.636{\scriptsize±0.051} & 0.614{\scriptsize±0.097}                & \textbf{0.710{\scriptsize±0.016}}                     & 0.624{\scriptsize±0.005}                     & {\color[HTML]{FF0000} \textbf{0.713{\scriptsize±0.003}}}                  \\ \cline{2-16} 
\multicolumn{1}{c|}{}                                       & \textbf{Mean}               & 0.674{\scriptsize±0.034}  & 0.466{\scriptsize±0.030} & 0.624{\scriptsize±0.024} & 0.646{\scriptsize±0.014}                &   \textbf{0.696{\scriptsize±0.011}}  & 0.660{\scriptsize±0.007}                    &   {\color[HTML]{FF0000} \textbf{0.707{\scriptsize±0.007}}}      & 0.724{\scriptsize±0.032}                     & 0.626{\scriptsize±0.041} & 0.635{\scriptsize±0.043} & 0.685{\scriptsize±0.016 }               & \textbf{0.733{\scriptsize±0.011}}                     & 0.709{\scriptsize±0.006}                     & {\color[HTML]{FF0000} \textbf{0.747{\scriptsize±0.002}}}                  \\ \hline
\multicolumn{1}{c|}{\multirow{3}{*}{\textbf{ELPV}}}         & \textbf{Mono}               & 0.563{\scriptsize±0.102}  & 0.649{\scriptsize±0.027} & 0.717{\scriptsize±0.025} & 0.620{\scriptsize±0.057}                &  \textbf{0.762{\scriptsize±0.017}}   & 0.638{\scriptsize±0.019 }                    &    {\color[HTML]{FF0000} \textbf{0.774{\scriptsize±0.013}} }      & 0.569{\scriptsize±0.035}                     & 0.756{\scriptsize±0.045} & 0.629{\scriptsize±0.072} & 0.639{\scriptsize±0.067}                & \textbf{0.735{\scriptsize±0.008}}                     & 0.663{\scriptsize±0.007}                     & {\color[HTML]{FF0000} \textbf{0.745{\scriptsize±0.004}} }                 \\
\multicolumn{1}{c|}{}                                       & \textbf{Poly}               & 0.665{\scriptsize±0.173}  & 0.483{\scriptsize±0.247} & 0.665{\scriptsize±0.021} & \textbf{0.705{\scriptsize±0.011}}                &    0.681{\scriptsize±0.026} & {\color[HTML]{FF0000} \textbf{0.717{\scriptsize±0.007}}}                     &    \textbf{0.705{\scriptsize±0.006}}       & 0.796{\scriptsize±0.084 }                    & 0.734{\scriptsize±0.078} & 0.662{\scriptsize±0.042} & 0.806{\scriptsize±0.027}                & 0.806{\scriptsize±0.004}                     & {\color[HTML]{FF0000} \textbf{0.842{\scriptsize±0.003}}}                     & \textbf{0.831{\scriptsize±0.011}}                  \\ \cline{2-16} 
    \multicolumn{1}{c|}{}                                       & \textbf{Mean}               & 0.614{\scriptsize±0.048}  & 0.566{\scriptsize±0.111} & 0.691{\scriptsize±0.008} & 0.663{\scriptsize±0.008}                &   \textbf{0.722{\scriptsize±0.006}}  & 0.678{\scriptsize±0.006}                     &   {\color[HTML]{FF0000} \textbf{0.740{\scriptsize±0.003}}}      & 0.683{\scriptsize±0.047}                     & 0.745{\scriptsize±0.020} & 0.646{\scriptsize±0.032} & 0.722{\scriptsize±0.018}                & \textbf{0.771{\scriptsize±0.005}}                     & 0.752{\scriptsize±0.005}                     & {\color[HTML]{FF0000} \textbf{0.788{\scriptsize±0.003}} }                 \\ \hline
\multicolumn{1}{c|}{\multirow{10}{*}{\textbf{Mastcam}}}     & \textbf{Bedrock}            & 0.636{\scriptsize±0.072}  & 0.532{\scriptsize±0.036} & 0.499{\scriptsize±0.056} & 0.508{\scriptsize±0.107}                &  \textbf{0.653{\scriptsize±0.019}}   & 0.533{\scriptsize±0.065 }                    &    {\color[HTML]{FF0000} \textbf{0.679{\scriptsize±0.012}} }     & 0.636{\scriptsize±0.068}                     & 0.512{\scriptsize±0.062} & 0.499{\scriptsize±0.098} & 0.586{\scriptsize±0.012}                & \textbf{0.654{\scriptsize±0.013}}                     & 0.589{\scriptsize±0.010}                     & {\color[HTML]{FF0000} \textbf{0.673{\scriptsize±0.006}} }                 \\
\multicolumn{1}{c|}{}                                       & \textbf{Broken-rock}        & {\color[HTML]{FF0000} \textbf{0.699{\scriptsize±0.058}}}  & 0.544{\scriptsize±0.088} & 0.569{\scriptsize±0.025} & 0.558{\scriptsize±0.016}                &   0.640{\scriptsize±0.023}  & 0.572{\scriptsize±0.014}                     &   \textbf{0.661{\scriptsize±0.009}}       & \textbf{0.712{\scriptsize±0.052}}                     & 0.651{\scriptsize±0.063} & 0.608{\scriptsize±0.085} & 0.562{\scriptsize±0.033}                & 0.704{\scriptsize±0.007}                     & 0.572{\scriptsize±0.024}                     & {\color[HTML]{FF0000} \textbf{0.722{\scriptsize±0.004} }}                 \\
\multicolumn{1}{c|}{}                                       & \textbf{Drill-hole}         & {\color[HTML]{FF0000} \textbf{0.697{\scriptsize±0.074}}}  & 0.636{\scriptsize±0.066} & 0.539{\scriptsize±0.077} & 0.555{\scriptsize±0.026}                &   0.642{\scriptsize±0.035}  & 0.563{\scriptsize±0.012}                     &    \textbf{0.654{\scriptsize±0.004}}      & 0.682{\scriptsize±0.042}                     & 0.660{\scriptsize±0.002} & 0.601{\scriptsize±0.009} & 0.590{\scriptsize±0.074}                & \textbf{0.757{\scriptsize±0.008}}                     & 0.610{\scriptsize±0.075 }                    & {\color[HTML]{FF0000} \textbf{0.760{\scriptsize±0.003}} }                 \\
\multicolumn{1}{c|}{}                                       & \textbf{Drt}                & {\color[HTML]{FF0000} \textbf{0.735{\scriptsize±0.020}}}  & 0.624{\scriptsize±0.042} & 0.591{\scriptsize±0.042} & 0.570{\scriptsize±0.048}                &   \textbf{0.733{\scriptsize±0.027}}  & 0.581{\scriptsize±0.023}                     &  0.724{\scriptsize±0.006}        & \textbf{0.761{\scriptsize±0.062}}                     & 0.616{\scriptsize±0.048} & 0.652{\scriptsize±0.024} & 0.620{\scriptsize±0.031}                & 0.757{\scriptsize±0.006}                     & 0.629{\scriptsize±0.016}                     & {\color[HTML]{FF0000} \textbf{0.772{\scriptsize±0.004}}}                  \\
\multicolumn{1}{c|}{}                                       & \textbf{Dump-pile}          & 0.682{\scriptsize±0.022}  & 0.545{\scriptsize±0.127} & 0.508{\scriptsize±0.021} & 0.510{\scriptsize±0.008}                &   \textbf{0.741{\scriptsize±0.022}}  & 0.519{\scriptsize±0.004 }                    &    {\color[HTML]{FF0000} \textbf{0.756{\scriptsize±0.011}}}      & 0.750{\scriptsize±0.037}                     & 0.696{\scriptsize±0.047} & 0.700{\scriptsize±0.070} & 0.689{\scriptsize±0.070}                & \textbf{0.757{\scriptsize±0.008}}                     & 0.695{\scriptsize±0.021}                     & {\color[HTML]{FF0000} \textbf{0.802{\scriptsize±0.005}}}                  \\
\multicolumn{1}{c|}{}                                       & \textbf{Float}              & {\color[HTML]{FF0000} \textbf{0.711{\scriptsize±0.041}}}  & 0.530{\scriptsize±0.075} & 0.551{\scriptsize±0.030} & 0.507{\scriptsize±0.039}                &  0.688{\scriptsize±0.031}   & 0.524{\scriptsize±0.017 }                    &    \textbf{0.702{\scriptsize±0.005}}      & 0.718{\scriptsize±0.064}                     & 0.671{\scriptsize±0.032} & 0.736{\scriptsize±0.041} & 0.640{\scriptsize±0.012}                & \textbf{0.749{\scriptsize±0.009}}                     & 0.647{\scriptsize±0.008 }                    & {\color[HTML]{FF0000} \textbf{0.765{\scriptsize±0.002}}}                  \\
\multicolumn{1}{c|}{}                                       & \textbf{Meteorite}          & {\color[HTML]{FF0000} \textbf{0.669{\scriptsize±0.037}}}  & 0.476{\scriptsize±0.014} & 0.462{\scriptsize±0.077} & 0.436{\scriptsize±0.033}                &   0.604{\scriptsize±0.020}  & 0.463{\scriptsize±0.008}                     &   \textbf{0.616{\scriptsize±0.013}}       & 0.647{\scriptsize±0.030 }                    & 0.473{\scriptsize±0.047} & 0.568{\scriptsize±0.053} & 0.561{\scriptsize±0.053}                & \textbf{0.689{\scriptsize±0.010}}                     & 0.572{\scriptsize±0.015}                     & {\color[HTML]{FF0000} \textbf{0.691{\scriptsize±0.001}}}                  \\
\multicolumn{1}{c|}{}                                       & \textbf{Scuff}              & {\color[HTML]{FF0000} \textbf{0.679{\scriptsize±0.048}}}  & 0.492{\scriptsize±0.037} & 0.508{\scriptsize±0.070} & 0.496{\scriptsize±0.121}                &  0.573{\scriptsize±0.017 }  & 0.515{\scriptsize±0.013}           &   \textbf{0.581{\scriptsize±0.020} }      & {\color[HTML]{FF0000} \textbf{0.676{\scriptsize±0.019}}}                     & 0.504{\scriptsize±0.052} & 0.575{\scriptsize±0.042} & 0.447{\scriptsize±0.043}                & 0.626{\scriptsize±0.016}                     & 0.506{\scriptsize±0.104}                     & \textbf{0.656{\scriptsize±0.009} }                 \\
\multicolumn{1}{c|}{}                                       & \textbf{Veins}              & {\color[HTML]{FF0000} \textbf{0.688{\scriptsize±0.069}}}  & 0.489{\scriptsize±0.028} & 0.493{\scriptsize±0.052} & 0.530{\scriptsize±0.006 }               &   0.650{\scriptsize±0.012}  & 0.548{\scriptsize±0.010 }                    &   \textbf{0.687{\scriptsize±0.017}}       & {\color[HTML]{FF0000} \textbf{0.686{\scriptsize±0.053}} }                    & 0.510{\scriptsize±0.090} & 0.608{\scriptsize±0.044} & 0.577{\scriptsize±0.029}                & 0.644{\scriptsize±0.007}                     & 0.598{\scriptsize±0.037}                     & \textbf{0.650{\scriptsize±0.003 }}                 \\ \cline{2-16} 
\multicolumn{1}{c|}{}                                       & \textbf{Mean}               & {\color[HTML]{FF0000} \textbf{0.689{\scriptsize±0.037}}}  & 0.541{\scriptsize±0.007} & 0.524{\scriptsize±0.013} & 0.519{\scriptsize±0.014}                &   0.658{\scriptsize±0.021}   & 0.535{\scriptsize±0.003}                     &    \textbf{0.673{\scriptsize±0.010}}      & 0.697{\scriptsize±0.014 }                    & 0.588{\scriptsize±0.016} & 0.616{\scriptsize±0.021} & 0.588{\scriptsize±0.025}                & \textbf{0.704{\scriptsize±0.007}}                     & 0.602{\scriptsize±0.008}                     & {\color[HTML]{FF0000} \textbf{0.721{\scriptsize±0.003}} }                 \\ \hline
\multicolumn{1}{c|}{\multirow{5}{*}{\textbf{Hyper-Kvasir}}} & \textbf{Barretts}           & 0.382{\scriptsize±0.117}  & 0.438{\scriptsize±0.111} & 0.703{\scriptsize±0.040} & 0.682{\scriptsize±0.007}                &  \textbf{0.788{\scriptsize±0.008}}   & 0.701{\scriptsize±0.013}                     &   {\color[HTML]{FF0000} \textbf{0.792{\scriptsize±0.007}}}       & 0.698{\scriptsize±0.037}                     & 0.540{\scriptsize±0.014} & 0.764{\scriptsize±0.066} & \textbf{0.837{\scriptsize±0.014}}                & 0.820{\scriptsize±0.005}                     & {\color[HTML]{FF0000} \textbf{0.850{\scriptsize±0.002}} }                    & 0.829{\scriptsize±0.002}                  \\
\multicolumn{1}{c|}{}                                       & \textbf{Barretts-short-seg} & 0.367{\scriptsize±0.050}  & 0.532{\scriptsize±0.075} & 0.538{\scriptsize±0.033} & 0.608{\scriptsize±0.077}                &   \textbf{0.643{\scriptsize±0.013}}  & 0.629{\scriptsize±0.027 }                    &   {\color[HTML]{FF0000} \textbf{0.651{\scriptsize±0.006 }}}     & 0.661{\scriptsize±0.034}                     & 0.480{\scriptsize±0.107} & 0.810{\scriptsize±0.034} & 0.790{\scriptsize±0.017}                & \textbf{0.829{\scriptsize±0.006}}                     & 0.812{\scriptsize±0.005}                     & {\color[HTML]{FF0000} \textbf{0.895{\scriptsize±0.003}} }                 \\
\multicolumn{1}{c|}{}                                       & \textbf{Esophagitis-a}      & 0.518{\scriptsize±0.063}  & 0.491{\scriptsize±0.084} & 0.536{\scriptsize±0.040} & 0.567{\scriptsize±0.034}                &  \textbf{0.759{\scriptsize±0.015} }  & 0.583{\scriptsize±0.015}                     &   {\color[HTML]{FF0000} \textbf{0.760{\scriptsize±0.006}}}       & 0.820{\scriptsize±0.034}                     & 0.646{\scriptsize±0.036} & 0.815{\scriptsize±0.022} & 0.867{\scriptsize±0.004}                & 0.854{\scriptsize±0.006}                     & \textbf{0.876{\scriptsize±0.002}}                     & {\color[HTML]{FF0000} \textbf{0.878{\scriptsize±0.021}}}                  \\
\multicolumn{1}{c|}{}                                       & \textbf{Esophagitis-b-d}    & 0.358{\scriptsize±0.039}  & 0.457{\scriptsize±0.086} & 0.505{\scriptsize±0.039} & 0.535{\scriptsize±0.025}                &  \textbf{0.604{\scriptsize±0.022}}   & 0.562{\scriptsize±0.010}                     &    {\color[HTML]{FF0000} \textbf{0.622{\scriptsize±0.014}}}      & 0.611{\scriptsize±0.017 }                    & 0.621{\scriptsize±0.042} & 0.754{\scriptsize±0.073} & 0.812{\scriptsize±0.025}                & 0.785{\scriptsize±0.006}                     & {\color[HTML]{FF0000} \textbf{0.842{\scriptsize±0.003}}}                     & \textbf{0.815{\scriptsize±0.010}}                  \\ \cline{2-16} 
\multicolumn{1}{c|}{}                                       & \textbf{Mean}               & 0.406{\scriptsize±0.018} & 0.480{\scriptsize±0.044} & 0.571{\scriptsize±0.004} & 0.598{\scriptsize±0.006}                &   \textbf{0.699{\scriptsize±0.009} } & 0.619{\scriptsize±0.005}                     &     {\color[HTML]{FF0000} \textbf{0.706{\scriptsize±0.007}}}     & 0.698{\scriptsize±0.021}                     & 0.571{\scriptsize±0.014} & 0.786{\scriptsize±0.021} & 0.827{\scriptsize±0.008}                & 0.822{\scriptsize±0.013}                     & \textbf{0.845{\scriptsize±0.003}}                     & {\color[HTML]{FF0000} \textbf{0.854{\scriptsize±0.004} }}                 \\ \hline
\end{tabular}}
\caption{AUC results(mean±std) on nine real-world AD datasets under the hard setting. Best results and the second-best results are respectively highlighted in {\color[HTML]{FF0000} \textbf{Red}} and \textbf{Bold}. Carpet and Meta\_nut are two subsets of MVTec AD. The same set of datasets is used as in \cite{ding2022catching}.}
\label{hard_detail}
\end{table*}

\begin{table}[h]
\resizebox{\linewidth}{!}{
\begin{tabular}{|cccccc|}
\hline
\multicolumn{1}{|c|}{\textbf{Datsset}}                       & \multicolumn{1}{c|}{\textbf{Subset}}             & \textbf{DRA}  & \textbf{+ RamHADG}    & \textbf{+ HADG} & \textbf{AHL(DevNet)}  \\ \hline
\multicolumn{1}{|c|}{\multirow{6}{*}{\textbf{Carpet}}}       & \multicolumn{1}{c|}{\textbf{Color}}              & 0.899{\scriptsize±0.019}          & 0.917{\scriptsize±0.004}          & \textbf{0.919{\scriptsize±0.003}}           & {\color[HTML]{FF0000} \textbf{0.929{\scriptsize±0.007}}}          \\
\multicolumn{1}{|c|}{}                                       & \multicolumn{1}{c|}{\textbf{Cut}}                & 0.942{\scriptsize±0.012}          & 0.938{\scriptsize±0.004}          & \textbf{0.943{\scriptsize±0.001}}           & {\color[HTML]{FF0000} \textbf{0.943{\scriptsize±0.002}}}          \\
\multicolumn{1}{|c|}{}                                       & \multicolumn{1}{c|}{\textbf{Hole}}               & \textbf{0.958{\scriptsize±0.031}}          & 0.954{\scriptsize±0.010}          & 0.952{\scriptsize±0.002}           & {\color[HTML]{FF0000} \textbf{0.960{\scriptsize±0.003}}}          \\
\multicolumn{1}{|c|}{}                                       & \multicolumn{1}{c|}{\textbf{Metal}}              & 0.916{\scriptsize±0.017}          & \textbf{0.919{\scriptsize±0.008}}          & 0.914{\scriptsize±0.006}           & {\color[HTML]{FF0000} \textbf{0.921{\scriptsize±0.003}}}          \\
\multicolumn{1}{|c|}{}                                       & \multicolumn{1}{c|}{\textbf{Thread}}             & 0.985{\scriptsize±0.005}          & 0.985{\scriptsize±0.003}          & \textbf{0.988{\scriptsize±0.002}}           & {\color[HTML]{FF0000} \textbf{0.991{\scriptsize±0.001}}}          \\
\multicolumn{1}{|c|}{}                                       & \multicolumn{1}{c|}{\textbf{Mean}}               & 0.940{\scriptsize±0.006}          & \textbf{0.943{\scriptsize±0.002}}          & \textbf{0.943{\scriptsize±0.003}}  & {\color[HTML]{FF0000} \textbf{0.949{\scriptsize±0.002}}}          \\ \hline
\multicolumn{1}{|c|}{\multirow{7}{*}{\textbf{AITEX}}}        & \multicolumn{1}{c|}{\textbf{Broken end}}         & 0.708{\scriptsize±0.062}          & 0.714{\scriptsize±0.011}          & \textbf{0.719{\scriptsize±0.005}}           & {\color[HTML]{FF0000} \textbf{0.735{\scriptsize±0.010}}}          \\
\multicolumn{1}{|c|}{}                                       & \multicolumn{1}{c|}{\textbf{Broken pick}}        & 0.671{\scriptsize±0.034}          & 0.670{\scriptsize±0.003}          & \textbf{0.678{\scriptsize±0.005}}           & {\color[HTML]{FF0000} \textbf{0.683{\scriptsize±0.002}}}          \\
\multicolumn{1}{|c|}{}                                       & \multicolumn{1}{c|}{\textbf{Cut selvage}}        & 0.777{\scriptsize±0.021}          & 0.777{\scriptsize±0.009}          & \textbf{0.779{\scriptsize±0.018}}           & {\color[HTML]{FF0000} \textbf{0.781{\scriptsize±0.006}}}          \\
\multicolumn{1}{|c|}{}                                       & \multicolumn{1}{c|}{\textbf{Fuzzyball}}          & 0.749{\scriptsize±0.033}          & 0.742{\scriptsize±0.010}          & \textbf{0.756{\scriptsize±0.021}}           & {\color[HTML]{FF0000} \textbf{0.775{\scriptsize±0.024}}}          \\
\multicolumn{1}{|c|}{}                                       & \multicolumn{1}{c|}{\textbf{Nep}}                & 0.784{\scriptsize±0.025}          & 0.786{\scriptsize±0.007}          & \textbf{0.788{\scriptsize±0.005}}           & {\color[HTML]{FF0000} \textbf{0.792{\scriptsize±0.007}}}          \\
\multicolumn{1}{|c|}{}                                       & \multicolumn{1}{c|}{\textbf{Weft crack}}         & 0.710{\scriptsize±0.016}          & 0.708{\scriptsize±0.003}          & \textbf{0.711{\scriptsize±0.003}}           & {\color[HTML]{FF0000} \textbf{0.713{\scriptsize±0.003}}}          \\
\multicolumn{1}{|c|}{}                                       & \multicolumn{1}{c|}{\textbf{Mean}}               & 0.733{\scriptsize±0.011}          & 0.733{\scriptsize±0.005}          & \textbf{0.739{\scriptsize±0.007}}           & {\color[HTML]{FF0000} \textbf{0.747{\scriptsize±0.002}}}          \\ \hline
\multicolumn{1}{|c|}{\multirow{3}{*}{\textbf{elpv}}}         & \multicolumn{1}{c|}{\textbf{Mono}}               & 0.735{\scriptsize±0.008}          & 0.738{\scriptsize±0.008}          & \textbf{0.739{\scriptsize±0.006}}           & {\color[HTML]{FF0000} \textbf{0.745{\scriptsize±0.004 }}}         \\
\multicolumn{1}{|c|}{}                                       & \multicolumn{1}{c|}{\textbf{Poly}}               & 0.806{\scriptsize±0.004}          & 0.809{\scriptsize±0.006}          & \textbf{0.817{\scriptsize±0.012}}           & {\color[HTML]{FF0000} \textbf{0.831{\scriptsize±0.011}}}          \\
\multicolumn{1}{|c|}{}                                       & \multicolumn{1}{c|}{\textbf{Mean}}               & 0.771{\scriptsize±0.005}          & 0.774{\scriptsize±0.004}          & \textbf{0.784{\scriptsize±0.004}}           & {\color[HTML]{FF0000} \textbf{0.788{\scriptsize±0.003 }}}         \\ \hline
\multicolumn{1}{|c|}{\multirow{5}{*}{\textbf{Hyper-Kvasir}}} & \multicolumn{1}{c|}{\textbf{Barretts}}           & 0.820{\scriptsize±0.005}          & 0.819{\scriptsize±0.006 }         & \textbf{0.822{\scriptsize±0.009}}           & {\color[HTML]{FF0000} \textbf{0.829{\scriptsize±0.002}}}          \\
\multicolumn{1}{|c|}{}                                       & \multicolumn{1}{c|}{\textbf{Barretts-short-seg}} & 0.829{\scriptsize±0.006}          & 0.864{\scriptsize±0.012}          & \textbf{0.887{\scriptsize±0.016}}           & {\color[HTML]{FF0000} \textbf{0.895{\scriptsize±0.003 } }}        \\
\multicolumn{1}{|c|}{}                                       & \multicolumn{1}{c|}{\textbf{Esophagitis-a}}      & 0.854{\scriptsize±0.006}          & 0.863{\scriptsize±0.000}          & \textbf{0.871{\scriptsize±0.005}}           & {\color[HTML]{FF0000} \textbf{0.878{\scriptsize±0.021}}}          \\
\multicolumn{1}{|c|}{}                                       & \multicolumn{1}{c|}{\textbf{Esophagitis-b-d}}    & 0.785{\scriptsize±0.006}          & 0.795{\scriptsize±0.005}          & \textbf{0.806{\scriptsize±0.004 } }         & {\color[HTML]{FF0000} \textbf{0.815{\scriptsize±0.010}}}         \\
\multicolumn{1}{|c|}{}                                       & \multicolumn{1}{c|}{\textbf{Mean}}               & 0.822{\scriptsize±0.013 }         & 0.835{\scriptsize±0.004 }         & \textbf{0.847{\scriptsize±0.008}}           & {\color[HTML]{FF0000} \textbf{0.854{\scriptsize±0.004}}}          \\ \hline
\end{tabular}}
\caption{Ablation study class-level results of \coolname and its three main variants under hard settings. Best results and the second-best results are respectively highlighted in {\color[HTML]{FF0000} \textbf{Red}} and \textbf{Bold}.}
\label{ab_detail}
\end{table}

\subsection{More Ablation Study Results}
\vspace{0.1cm}
\noindent\textbf{Class-level Results under Hard Setting}
To evaluate the effectiveness of each module in our \coolname approach, we compared it with three variants: the base model (\textbf{DRA}), the base model with HADG component (\textbf{+ HADG}), and the base model using random heterogeneous anomaly distribution generation (\textbf{+ RamHADG}). Table~\ref{ab_detail} shows the detailed results of class-level anomaly detection in the hard setting. The results show that all the modules in the \coolname framework contribute to improving the detection performance on unseen anomaly classes, demonstrating the importance of anomaly heterogeneity. 

\begin{figure}[h]
    \centering
    \includegraphics[width=0.85\linewidth]{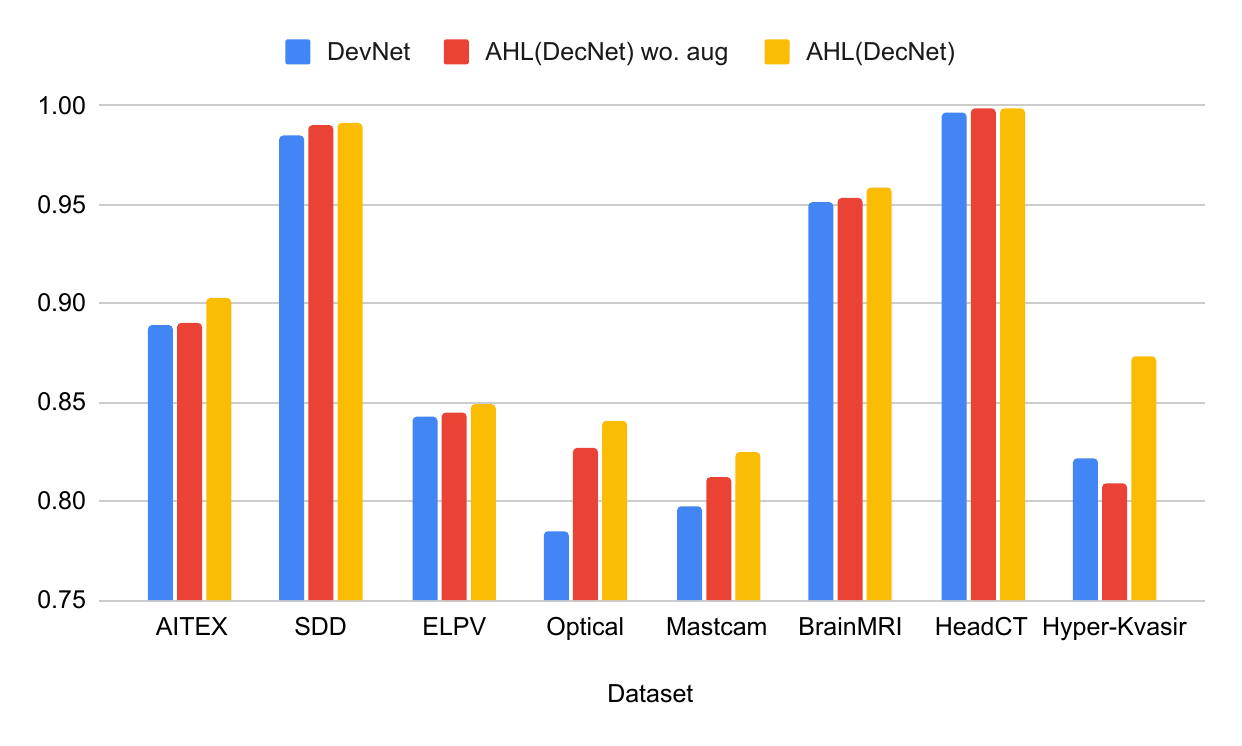}
    \caption{Comparison of the AUC performance between DevNet and \coolname(DRA) without the pseudo anomaly augmentation generation module. Here $wo.\ aug$ indicates the augmentation techniques are excluded during training.}
    \label{fig:aug}
\end{figure}

\vspace{0.1cm}
\noindent\textbf{Importance of Pseudo Anomalies}
Moreover, since DevNet~\cite{pang2021explainable} does not use pseudo anomalies, we also evaluate the impact of removing the pseudo anomalies on the \coolname framework when using DevNet as the base model. As shown in Figure~\ref{fig:aug}, \coolname(DevNet) remains substantially better than the original DevNet model in such cases.

\vspace{0.1cm}
\noindent\textbf{Effectiveness of Overlapping in HADG Module}
Table \ref{overlapping} shows the results of \coolname(DRA) using three overlapping types: overlap occurring exclusively in (1) the query set ($\mathcal{D}^s$-\textgreater{}$\mathcal{D}^q$), (2) the support set ($\mathcal{D}^q$-\textgreater{}$\mathcal{D}^s$), and (3)  in both sets ($\mathcal{D}^s<>\mathcal{D}^q$). 
It is clear that the overlapping between $\mathcal{D}^q$ and $\mathcal{D}^s$ typically does not lead to performance improvement. 
\begin{table}[ht]
\centering
\resizebox{0.8\linewidth}{!}{
\begin{tabular}{cc|lllclcl|lllclcc}
\hline
\textbf{Dataset}      & \textbf{$\mathcal{D}^s$-\textgreater{}$\mathcal{D}^q$} & \textbf{$\mathcal{D}^q$-\textgreater{}$\mathcal{D}^s$} & \textbf{$\mathcal{D}^s<>\mathcal{D}^q$} & \textbf{Ours} \\ \hline
\textbf{AITEX}        & 0.918                              & 0.917                              & 0.917                   & \textbf{0.925}      \\
\textbf{SDD}          & 0.988                              & 0.990                              & \textbf{0.991}          & \textbf{0.991}      \\
\textbf{ELPV}         & 0.853                              & 0.844                              & 0.842                   & \textbf{0.850}      \\
\textbf{BrainMRI}     & 0.970                              & 0.969                              & 0.970                   & \textbf{0.977}      \\
\textbf{HeadCT}       & 0.993                              & 0.991                              & 0.993                   & \textbf{0.993}      \\
\textbf{Hyper-Kvasir} & 0.872                              & 0.875                              & 0.876                   & \textbf{0.880}      \\ \hline
\end{tabular}}
\vspace{-0.3cm}
\captionsetup{font={footnotesize}}
\caption{\textit{Results with various overlaps.}}
\label{overlapping}
\vspace{-0.2cm}
\end{table}

\end{document}